\documentclass[10pt,twocolumn,letterpaper]{article}

\usepackage[table]{xcolor}
\usepackage{tikz}
\usepackage{cvpr}
\usepackage{times}
\usepackage{epsfig}
\usepackage{graphicx}
\usepackage{amsmath}
\usepackage{amssymb}
\usepackage{subfiles}
\usepackage{adjustbox}
\usepackage{paralist}

\usepackage{pgfplots} 
\usepackage{pgfplotstable} 
\pgfplotsset{compat=newest} 
\usetikzlibrary{plotmarks} 
\usepackage[table]{xcolor}

\usepackage{authblk}

\usepackage[pagebackref=true,breaklinks=true,letterpaper=true,colorlinks,bookmarks=false]{hyperref}

\cvprfinalcopy 

\def\cvprPaperID{2680} 

\ifcvprfinal\fi
\begin{document}

\title{Iterative Deep Learning for Network Topology Extraction}

\author[1]{C. Ventura}
\author[2]{J. Pont-Tuset}
\author[2]{S. Caelles}
\author[2]{K.-K. Maninis}
\author[2]{L. Van Gool}
\affil[1]{Universitat Oberta de Catalunya, Barcelona, Spain}
\affil[2]{Computer Vision Lab, ETH Z\"{u}rich, Switzerland}

\maketitle

\begin{abstract}
This paper tackles the task of estimating the topology of filamentary networks such as retinal vessels and road networks.
Building on top of a global model that performs a dense semantical classification of the pixels of the image, we design a Convolutional Neural Network (CNN) that predicts the local connectivity between the central pixel of an input patch and its border points.
By iterating this local connectivity we sweep the whole image and infer the global topology of the filamentary network, inspired by a human delineating a complex network with the tip of their finger.

We perform an extensive and comprehensive qualitative and quantitative evaluation on two tasks: retinal veins and arteries topology extraction and road network estimation.
In both cases, represented by two publicly available datasets (DRIVE and Massachusetts Roads), we show superior performance to very strong baselines.
\end{abstract}


\section{Introduction}
Deep learning has gone a long way since its jump to fame in the field of computer vision thanks to the outstanding results in the Imagenet~\cite{Rus+15} image classification competition back in 2012~\cite{Krizhevsky2012}.
We have witnessed the appearance of deeper~\cite{SiZi15} and deeper~\cite{He+16} architectures and the generalization to object detection with the well-known trilogy of R-CNNs~\cite{Gir+14, Gir15, Ren+15}.
Convolutional Neural Networks (CNNs) have played a central role in this development.

A significant step forward was done with the introduction of CNNs for dense prediction, in which the output of the system was not a classification of an image or bounding box into certain categories, but each pixel would receive an output decision.
The seminal fully convolutional networks~\cite{LSD15} was able to perform per-pixel semantic segmentation thanks to an architecture without fully connected layers (i.e.\ fully convolutional).
Many tasks have been tackled from this perspective since then: semantic instance segmentation~\cite{Li+17, He+17}, edge detection~\cite{XiTu17}, medical image segmentation~\cite{maninis2016deep}, etc.

Other tasks, however, have a richer output structure beyond a per-pixel classification, and a higher abstraction of the result is expected.
Notable examples that have already been tackled by CNNs are the estimation of the human pose~\cite{NYD16}, or the room layout~\cite{Lee+17} from an image.
The common denominator of these tasks is that one expects an abstracted model of the result rather than a set of pixel classifications.

This work falls into this category by bringing the power of CNNs to the estimation of the \textbf{topology of filamentary networks} such as retinal vessels and road networks.
The structured output is of critical importance and priceless value in these applications: rather than knowing exactly which pixels in a satellite image are road or not, detecting whether two points are connected and how is arguably more informative.
In the medical field, knowing which is the widest and straightest vessel that is connected to an obstructed point helps doctors apply the needed cure more effectively.

\begin{figure}[t]
\centering
\includegraphics[width=0.96\linewidth]{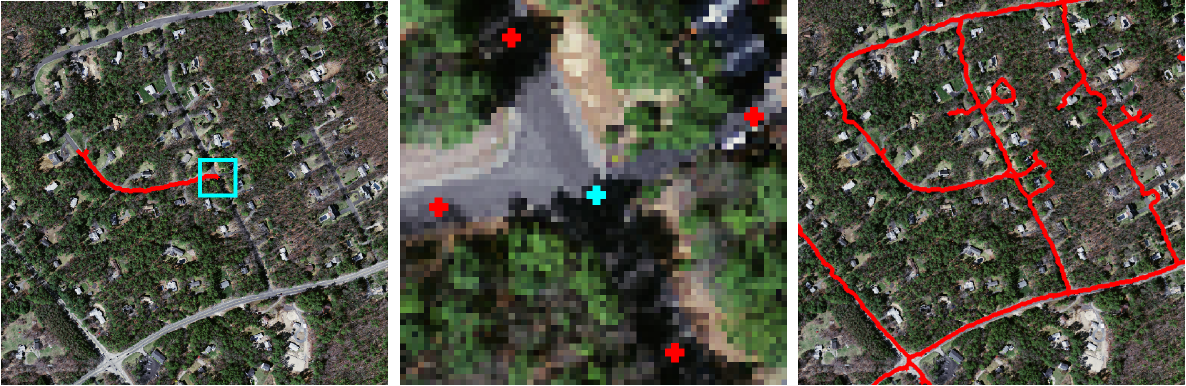}
\caption{Patch-based iterative approach for network topology extraction. \emph{Left}: Current state at some point of the iterative approach where the patch-level model for connectivity is applied over the
blue square. \emph{Center}: Detections at the local patch for the points at the border (in red) connected to the central point (in blue). \emph{Right}: Final result once the 
iterative approach ends.}
\label{fig:teaser-img}
\end{figure}

If one thinks how humans would extract the topology of an entangled graph network from an image, it might quickly come to mind the image of them tracing the filaments with the finger and \textit{sweeping} the connected paths continuously.
Inspired by this, we propose an iterative deep learning approach that sequentially connects dots within the filaments until it \textit{sweeps} all the visible network.

More specifically, we train a CNN on small patches that localizes input and exit points of the filaments within the patch (see image in the middle from Figure~\ref{fig:teaser-img}). 
By iteratively connecting these dots we obtain the global topology (graph) of the network (see right image from Figure~\ref{fig:teaser-img}).
We tackle the extraction of the topology of retinal vessels (veins and arteries) from fundus images and of road networks from aerial photos.
We experiment on a variety of datasets to show that our algorithm improves over some very strong baselines and provides accurate representations of the topology of both cases.

\section{Related Work}
\paragraph{Curvilinear Structure Segmentation and Tracing:}
Tracing of curvilinear structures has been of broad interest in a range of applications, varying from blood vessel segmentation, roadmap segmentation, and reconstruction of human vasculature. Hessian-based methods rely on derivatives, to guide the development of a snake~\cite{WAC11}, or to detect vessel boundaries~\cite{Ban+12}. Model-based methods rely on strong assumptions about the geometric shapes of the filamentary structures~\cite{LaCh08, Soa+06}. Learning-based methods emerged for the task, using support vector machines on line operators~\cite{RiPe07}, fully-connected CRFs~\cite{OrBl14}, gradient-boosting~\cite{Bec+13}, classification trees~\cite{GuCh15}, or nearest neighbours~\cite{SLF15}. Closer to our approach, the most recent methods rely on Fully Convolutional Neural Networks (FCNs), to segment retinal blood vessels~\cite{maninis2016deep, Fu+16} , or recover vascular boundaries~\cite{Mer+16}. Different than all the aforementioned method that result in binary structure maps, our method employs deep learning to trace the entire structure of the curvilinear structures, recovering their entire connectivity map. Also related to our method, the authors of~\cite{cheng2014tracing} trace blood vessels using directed graph theory. To the best of our knowledge, we are the first to apply deep learning for tracing curvilinear structures.

\paragraph{Road Centerline Detection:} 
Centerline detection has also followed the trend of curvilinear structure segmentation, with early attempts on gradient-based methods getting outperformed when stronger machine learning techniques emerged~\cite{WMS13, SLF14}. Sironi et al.~\cite{SLF15} model the relationships between neighbouring patches to reach the decision for the centerlines. Most recent works employ deep learning techniques, and include results on the Torontocity dataset~\cite{wang2017torontocity} (which has not been publicly released yet).~\cite{mattyus2017deep} is the most recent work on extracting the road topology from aerial images, and proposes a post processing 
algorithm that reasons about missing connections in the extracted road topology from an initial segmentation. In contrast, in our paper we propose an approach that learns the connectivity
of the roads at a local scale and is iteratively extended to the entire road network without relying on the results of an initial segmentation.

\section{Our Approach}

This section presents our approach, which combines a global scale for curvilinear structure segmentation and a local scale to estimate its connectivity. 
The current best approaches for curvilinear structure segmentation applies state-of-the-art deep learning techniques to obtain a segmentation map where each pixel is classified as 
belonging to the structure (foreground) or not (background).
The most relevant examples of such approaches are the VGG-based architecture used in DRIU~\cite{maninis2016deep} for vessel segmentation or the ResNet-based 
architecture used in~\cite{mattyus2017deep} for road segmentation.
Despite their good performance in segmentation evaluation measures,
one of the main drawbacks of these approaches is that they do not take any structure information into account.
In particular, these methods are blind to connectivity information among the points that lie in their predicted mask, since all points are assigned only a binary label.

\begin{figure}[t]
\centering
\setlength{\fboxsep}{0pt}
\fbox{\adjincludegraphics[width=0.48\linewidth,trim={{.03\width} {.03\height} {.03\width} {.03\height}},clip]{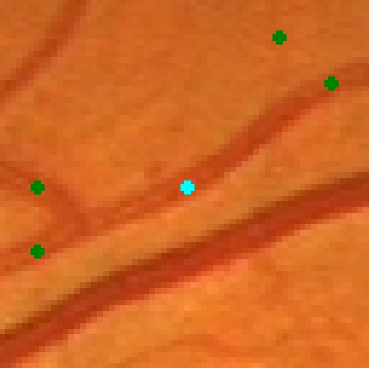}}
\fbox{\adjincludegraphics[width=0.48\linewidth,trim={{.03\width} {.03\height} {.03\width} {.03\height}},clip]{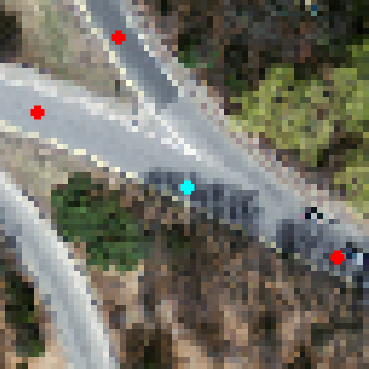}}
\caption{Examples of training patches for connectivity. The green/red points represent the locations from the patch border connected with the vessel or road indicated by the blue point 
in the center of the patch.}
\label{fig:training_patches}
\end{figure}

Section~\ref{sec:patch-level-connectivity} proposes a method that learns the connectivity of the elements at a local scale.
Given a patch of the image centered on a curvilinear structure, the model predicts the locations at the patch border connected with the centered structure.
Figure~\ref{fig:training_patches} shows some examples of how we formulate the local connectivity for retinal images and aerial images: we learn to predict the points on the border of the patch (green/red dots) that are connected to the center pixel (blue dot).

Once the local connectivity model is learned, it is iteratively applied to the image, connecting previous predictions with next ones, and gradually extracting the topology of the graph network, as explained in Section~\ref{sec:patch-level-iterative}.

We present our evaluation metrics in Section~\ref{sec:evaluation}.


\subsection{Patch-level learning for connectivity}
\label{sec:patch-level-connectivity}
As introduced above, the goal is to train a model to estimate the local connectivity in patches. 
The concept of connectivity is not a property from single points but from pairs of pixels.
Current architectures, however, are designed to estimate per-pixel properties rather than pairwise information. 
To solve this issue, the local network is designed to estimate which points in a patch are connected to a given input point.
Given a patch, therefore, we need to encode the position of the input and output points. 

In the context of human pose estimation~\cite{NYD16}, for instance, points have been encoded as heatmaps with Gaussians centered on them.

We follow the same approach and thus the output of our model is a per-pixel probability of being a connected point.
Instead of encoding the input point by adding an extra input channel with a heatmap marking its position, we follow a simpler approach.
We always place the input point the center of the patch, thus avoiding the extra input channel and further simplifying the model.
We see in the experiments that the model is indeed capable of learning that the central point is the input location we are interested in.

More precisely, we take the architecture of stacked hourglass networks~\cite{NYD16} (also used for human pose estimation) to learn the patch-based model for connectivity.
The network is trained using a set of $k\!\times\!k$-pixel patches from the training set with the pixel at the center of the patch belonging to the foreground (e.g. a vessel, a road, etc.).
The output is a heatmap that predicts the probability of each location being connected to the central point of the patch.

\begin{figure}
\centering
\setlength{\fboxsep}{0pt}
\fbox{\adjincludegraphics[width=0.48\linewidth,trim={{.03\width} {.03\height} {.03\width} {.03\height}},clip]{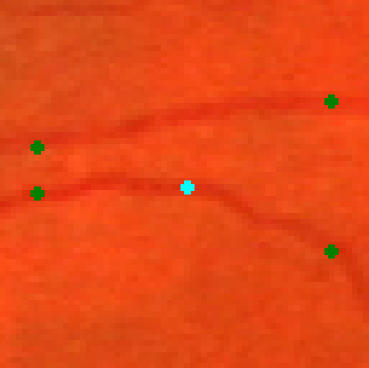}}\hfill
\setlength{\fboxsep}{0pt}
\fbox{\adjincludegraphics[width=0.48\linewidth,trim={{.03\width} {.03\height} {.03\width} {.03\height}},clip]{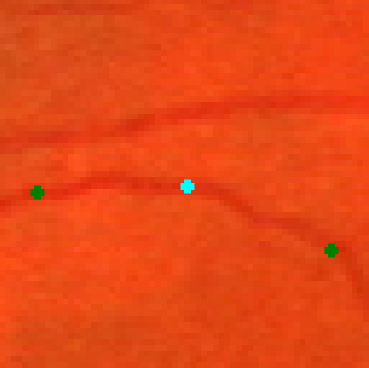}}\\[5pt]
\setlength{\fboxsep}{0pt}
\fbox{\adjincludegraphics[width=0.48\linewidth,trim={{.03\width} {.03\height} {.03\width} {.03\height}},clip]{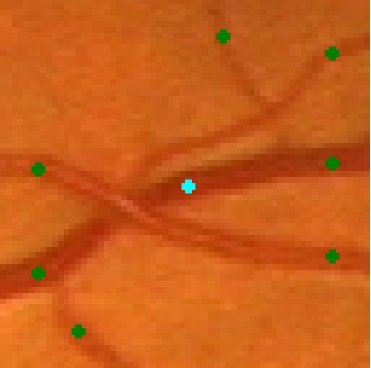}}\hfill
\setlength{\fboxsep}{0pt}
\fbox{\adjincludegraphics[width=0.48\linewidth,trim={{.03\width} {.03\height} {.03\width} {.03\height}},clip]{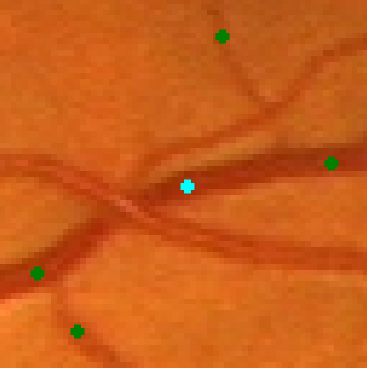}}
\caption{Examples of training patches for connectivity. The green points represent the locations from the patch border connected with the vessel indicated by the blue point 
in the center of the patch. The two patches from the first row show the difference between considering connectivity or not. The two patches from the second row show the difference between considering the type of vessel besides the connectivity.}
\label{fig:training_patches_different_modalities}
\end{figure}

In the case of the retinal images, the model is also trained to differentiate between the two types of vessel (arteries or veins), so the model is forced to learn not only the connectivity 
but also an artery-vein classification problem.
To illustrate this idea, Figure~\ref{fig:training_patches_different_modalities} shows some examples of connectivity for retinal images where we differentiate three
types of models. The first row compares two patches where all vessels that intersect the border patch have been marked (left), versus the ones that are connected to the vessel at the center of the patch (right).
The second row illustrates the difference between detecting the connectivity over any type of vessel (left), or forcing the connectivity to be over the same type of vessel (vein or artery).


\begin{figure}
\centering
\setlength{\fboxsep}{0pt}
\fbox{\adjincludegraphics[width=0.48\linewidth,trim={{.03\width} {.03\height} {.03\width} {.03\height}},clip]{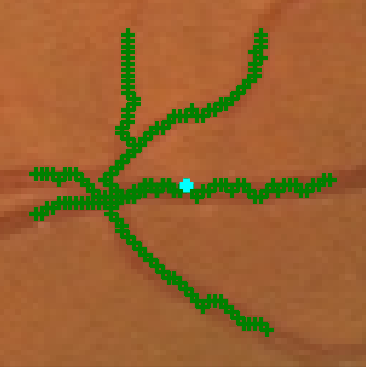}}\hfill
\setlength{\fboxsep}{0pt}
\fbox{\adjincludegraphics[width=0.48\linewidth,trim={{.03\width} {.03\height} {.03\width} {.03\height}},clip]{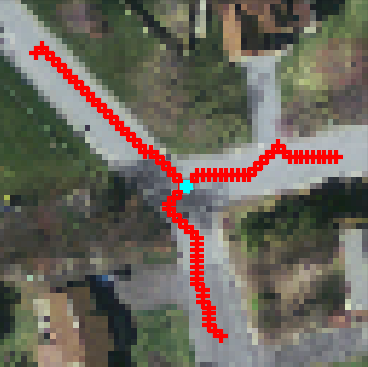}}\\[5pt]
\caption{Shortest path on semantic segmentation to connect the locations detected at the patch border with the patch center. }
\label{fig:example_connect}
\end{figure}

We finally connect the border locations to the center locations by computing the shortest path through the semantic segmentation computed from the global model introduced before, as shown in Figure~\ref{fig:example_connect}.
Note that the patch is local enough that a shortest path on the global model is reliable.

\subsection{Iterative delineation}
\label{sec:patch-level-iterative}
Once the patch-level model for connectivity has been learned, the model is applied iteratively through the image in order to extract the topology of the network, as a human delineating an image with its fingers not to lose the track.
We start from the point with highest foreground probability, given by the global model, as the starting point for the iterative sweeping approach.
We then center a patch on this point and find the set of locations at the border of the patch that are connected to the center, with their respective confidence values, using the local patch model.

We discard the locations with a confidence value below a certain threshold and add the remaining ones to a bag of points to be explored $B_E$.
For each predicted point, we store its location, its confidence value and its precedent predicted point (i.e. the point that was on the center of the patch when the point was predicted).
The predicted point $p$ from $B_E$ with the highest confidence value is removed from $B_E$ and inserted to a list of visited points $B_V$.

Then, $p$ is connected to its precedent predicted point using the Dijkstra~\cite{DIJKSTRA1959} algorithm over the segmentation probability map over the patch to find the minimum path between them.

We then iterate the process with a patch centered on $p$ and the new predicted points over the confidence threshold are appended to $B_E$ where they will \textit{compete} against the previous points in $B_E$ to be the next point to be explored.
This process is iteratively applied until $B_E$ is empty.
Note that the list of visited points $B_V$ is used to discard any point already explored and, therefore, to avoid revisiting the same points over and over again. 

Since in retinal images all vessels are connected through the optical disk, any vessel point from the image is reachable from any starting point used in the iterative approach. However, this is
not the case for any network image. For instance, aerial road images may content roads that are not connected between them. The same could also happen for a cropped retinal image where 
the entire retina is not visible and, therefore, there could be vessels not reachable from a single starting point. To prevent that some part of the network topology may have not been
extracted due to these missing connections, we select a new starting point for a new exploration once the previous $B_E$ is empty. We impose two constraints on the eligibility
for a new starting point: $(i)$ they have to be at a minimum distance of the areas already explored and $(ii)$ their confidence value on the segmentation probability map has to be over a 
minimum confidence threshold.
The iterative approach ends when there are no remaining points eligible for new starting points.

\subsection{Topology evaluation}
\label{sec:evaluation}
The output of our algorithm is a graph defining the topology of the input network, so we need metrics to evaluate their correctness.
We propose two different measures for this: a \textit{classical} precision-recall measure that evaluates which locations of the network are detected, and a metric to evaluate connectivity, by quantifying how many pairs of points are correctly or incorrectly connected.

To compute the classical precision-recall curve between two graphs, we build an image with a pixel-wide line sweeping all edges of the given graphs. We then apply the original precision-recall for boundaries \cite{martin2004learning} on these pair of images.
Precision $P$ refers to the ratio between the number of pixels correctly detected as boundary (true positives) and the number of pixels detected as boundary (true positives + false positives).
Recall $R$ refers to the ratio between the number of pixels correctly detected as boundary (true positives) and the number of pixels annotated as boundary in the ground truth (true positive + false negative).
We take the F measure between $P$ and $R$ as a trade-off metric.

\begin{figure}[t]
\centering
\setlength{\fboxsep}{0pt}
\fbox{\includegraphics[height=0.23\linewidth]{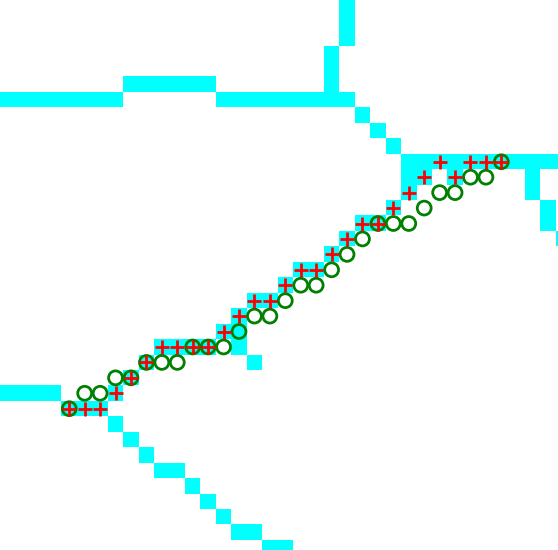}}\hfill
\fbox{\includegraphics[height=0.23\linewidth]{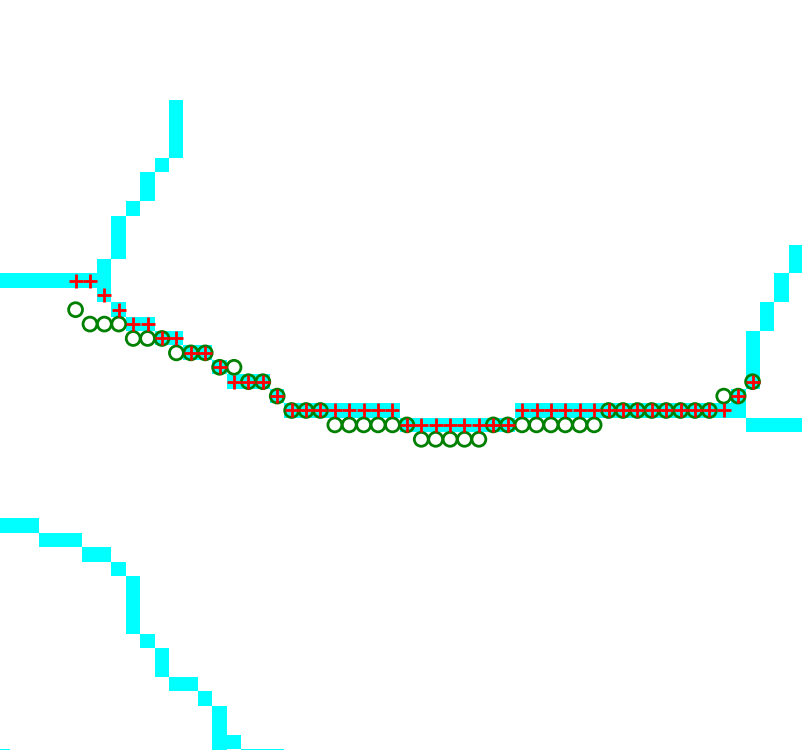}}\hfill
\fbox{\includegraphics[height=0.23\linewidth]{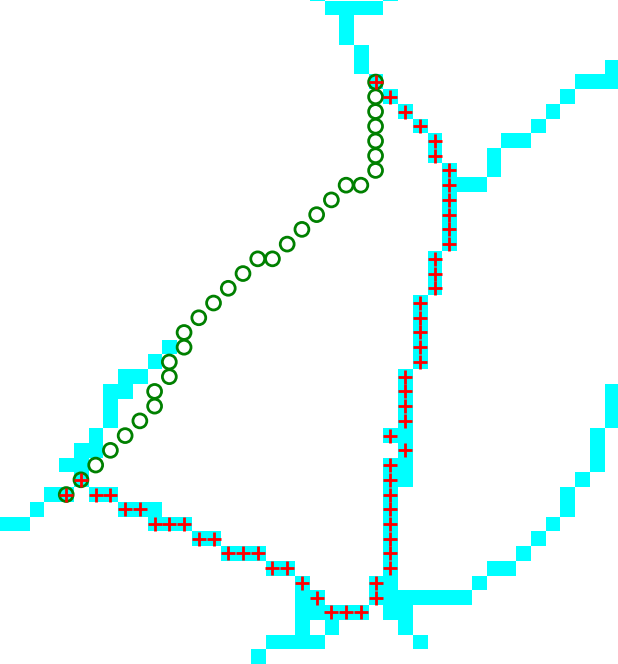}}\hfill
\fbox{\includegraphics[height=0.23\linewidth]{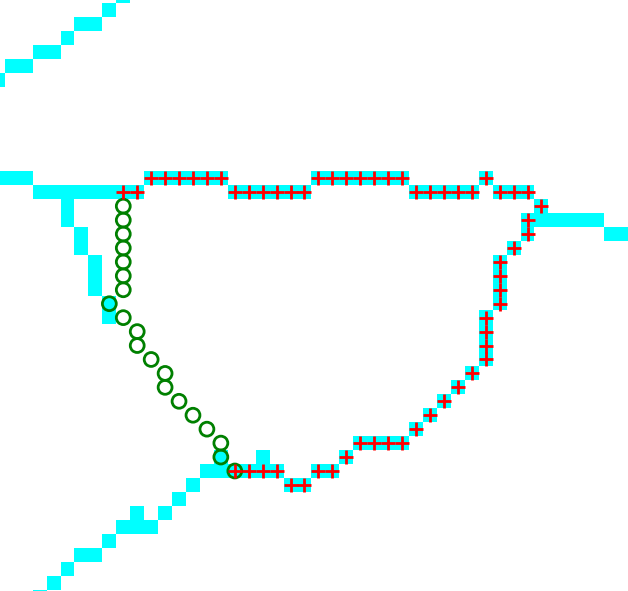}}
\caption{Examples of good (on the left) and bad (on the right) connectivity. Green pixels represent ground truth connections, blue pixels represent
predicted vessels with our iterative approach and red pixels represent the path found through predicted network.}
\label{fig:connectivity}
\end{figure}

The second measure is the connectivity $C$, inspired by the definition in~\cite{mattyus2017deep} as the ratio of segments which were estimated without discontinuities.
We define a segment in the graph as the curvilinear structure that connects two consecutive junctions in the ground-truth annotations, as well as connecting an endpoint 
and its closest connected junction (junctions refer to both crossovers and bifurcations). 
Two junctions are considered consecutive if there is no other junction within the line that connects them.
Figure~\ref{fig:connectivity} illustrates some examples of good and bad connectivity. Given the ground truth path between two consecutive junctions (showed in green) $p_{gt}$, the nearest point 
from the predicted network to each junction is retrieved. Then, the shortest path through the predicted network connecting the retrieved pair of points is computed (showed in red), which is 
referred to as $p_{pred}$. The ratio between the length of $p_{gt}$ and the length of $p_{pred}$ is computed. If the ratio is greater than 0.8 we consider that the ground truth path $p_{gt}$
has been estimated without discontinuities. In Figure~\ref{fig:connectivity}, the two images on the left show examples where the ground truth segment have been estimated without discontinuities,
whereas the two examples on the right are considered as not connected segments on the connectivity measure.

We propose to also have an F measure that combines precision $P$ with connectivity $C$. The reason is that a high connectivity $C$ value does not implies a good graph that defines the 
topology of the network. Whereas the connectivity measures the ratio of estimated segments without discontinuities, the precision measures how good the predicted locations along the segments are. 
Furthermore, the connectivity differs from the recall measure in the fact that the connectivity takes into account the distribution of the missing detections in the network. The connectivity
has a key role in the evaluation of the predicted topology since it is much worse having $k$ false negatives (missing detections) distributed along $k$ segments (worst scenario) than having the 
$k$ false negatives on a single segment (best scenario). Both previous scenarios have the same recall measure.

For the rest of the paper, $F^R$ stands for the F measure computed with recall and precision for boundaries values, whereas
$F^C$ stands for the $F$ measure computed between connectivity and precision.


%


\section{Experiments}

The experiments have been carried out in two different datasets with images that capture networks of curvilinear structures in two completely different contexts: the problem of blood vessel segmentation from eye fundus images, and the road segmentation from aerial images.
In both cases, our work aims at extracting the topology of the network preserving its
connectivity. 

\subsection{Vessel topology on retinal images}
\label{sec:experiments-vessels}

The experiments for retinal images have been carried out on the DRIVE~\cite{staal04ridge} dataset, which includes 40 eye fundus images and contains manual segmentation of the blood vessels by expert annotators.
We also take advantage of the work carried out by~\cite{estrada2015retinal}, which includes annotations as networks of linear segments and each linear segment is labelled as an artery or a vein.

\begin{figure}[t]
\centering
\setlength{\fboxsep}{0pt}
\fbox{\adjincludegraphics[width=0.32\linewidth,trim={{.05\width} {.04\height} {.05\width} {.05\height}},clip]{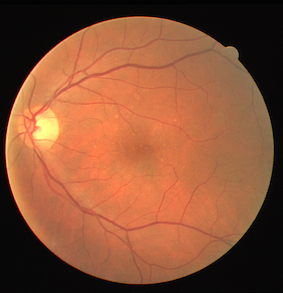}}\hfill
\fbox{\adjincludegraphics[width=0.32\linewidth,trim={{.05\width} {.04\height} {.05\width} {.05\height}},clip]{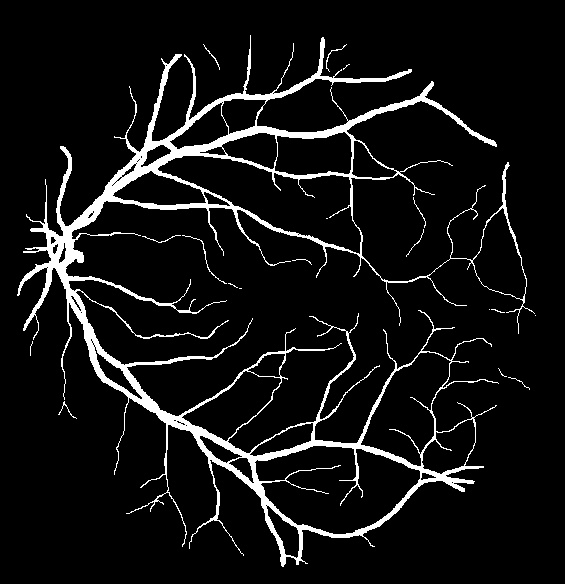}}\hfill
\fbox{\adjincludegraphics[width=0.32\linewidth,trim={{.027\width} {.01\height} {.01\width} {.01\height}},clip]{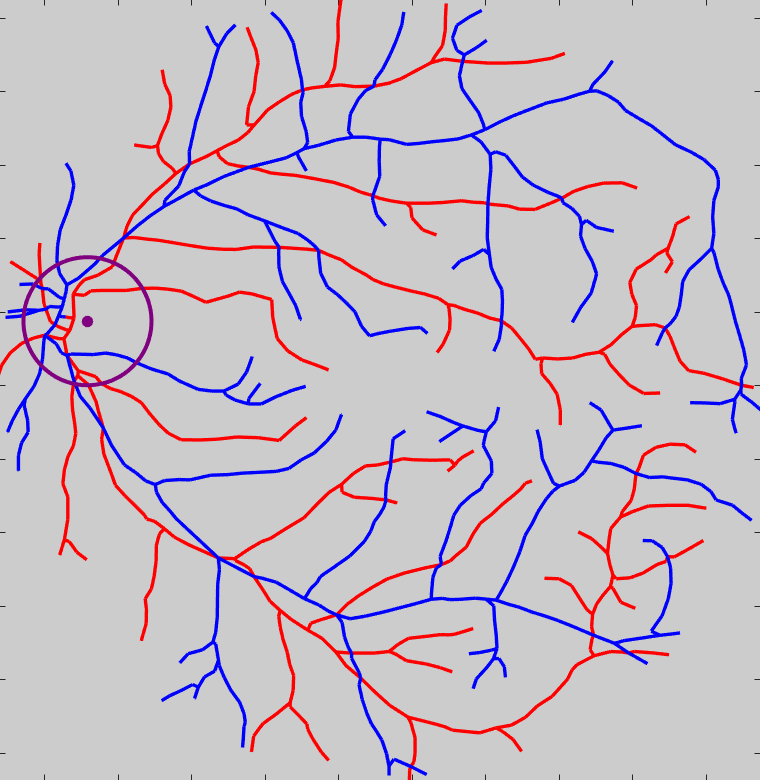}}\hfill
\caption{Example of an eye fundus image (left) and its available annotations (vessel segmentation on the middle, artery-vein network on the right).}
\label{fig:vessel_dataset}
\end{figure}

Figure~\ref{fig:vessel_dataset} shows one training image from DRIVE and its available annotations.
As a global model for segmentation, we use DRIU~\cite{maninis2016deep}, which is the state of the art for retinal vessel segmentation.

\paragraph{Patch-level evaluation:}
To train the patch-level model for connectivity we randomly select 50 patches with size 64$\times$64 pixels from each image of the training set, all of them centered on one of vertices of the graph annotations provided by~\cite{estrada2015retinal}.
The ground-truth locations for the connectivity at the patch level are found by intersecting the vessels with a square of side $s$ pixels (slightly smaller than the patch size) centered on the patch.
The ground-truth output heatmap is then generated by adding some Gaussian peaks centered in a subset of the found locations, depending on the configuration:

\begin{asparaitem}
 \item For the \emph{non-connectivity} model (e.g. top left in Figure~\ref{fig:training_patches_different_modalities}), all the intersection points are considered.
 \item For the \emph{connectivity} model, only the intersection points connected to the patch center along a way completely included in the patch are considered (e.g. top right or bottom left in Figure~\ref{fig:training_patches_different_modalities}).
 \item For the \emph{connectivity-av} model, only the intersection points connected to the patch center and belonging to the same type of vessel (artery or vein) as the vessel centered on the patch are considered (e.g. bottom right in Figure~\ref{fig:training_patches_different_modalities}).
\end{asparaitem}

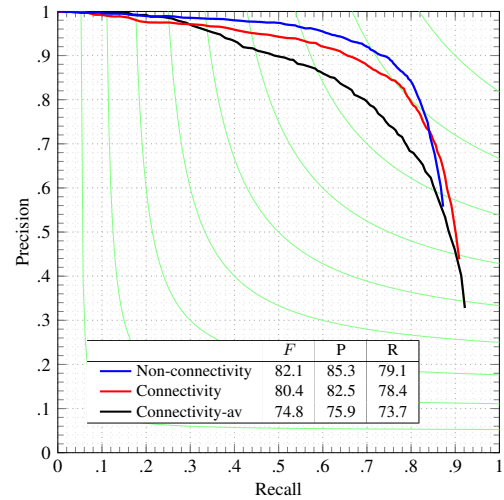
\begin{figure}[h]
\centering
\resizebox{0.8\linewidth}{!}{%
\begin{tikzpicture}[/pgfplots/width=1.1\linewidth, /pgfplots/height=1.1\linewidth]
    \begin{axis}[
                 ymin=0,ymax=1,xmin=0,xmax=1,
        		 xlabel=Recall,
        		 ylabel=Precision,
         		 xlabel shift={-2pt},
        		 ylabel shift={-3pt},
		         font=\small,
		         axis equal image=true,
		         enlargelimits=false,
		         clip=true,
        	     grid style=dotted, grid=both,
                 major grid style={white!65!black},
        		 minor grid style={white!85!black},
		 		 xtick={0,0.1,...,1.1},
        		 ytick={0,0.1,...,1.1},
         		 minor xtick={0,0.02,...,1},
		         minor ytick={0,0.02,...,1},
		         xticklabels={0,.1,.2,.3,.4,.5,.6,.7,.8,.9,1},
		         yticklabels={0,.1,.2,.3,.4,.5,.6,.7,.8,.9,1}]
        
    \foreach \f in {0.1,0.2,...,0.9}{%
       \addplot[white!50!green,line width=0.2pt,domain=(\f/(2-\f)):1,samples=200,forget plot]{(\f*x)/(2*x-\f)};
    }
    \addplot+[black,solid,mark=none, line width=1.1,forget plot] table[x=R,y=P] {PR_connected_same_vessel.txt};
    \label{fig:PR:samevessel}
    \addplot+[red,solid,mark=none, line width=1.1,forget plot] table[x=R,y=P] {PR_connected.txt};
    \label{fig:PR:connected}
    \addplot+[blue,solid,mark=none, line width=1.1,forget plot] table[x=R,y=P] {PR_not_connected.txt};
    \label{fig:PR:nonconnected}
    \node[anchor=south west] at (axis cs:0.05,0.05){\scalebox{0.9}{\begin{tabular}{ |l|c|c|c| } 
 \hline
 \rowcolor{white}& $F$ & P & R \\ 
 \hline
 \rowcolor{white}\ref{fig:PR:nonconnected} Non-connectivity & 82.1 & 85.3 & 79.1 \\ 
 \rowcolor{white}\ref{fig:PR:connected} Connectivity & 80.4 & 82.5 & 78.4 \\ 
 \rowcolor{white}\ref{fig:PR:samevessel} Connectivity-av & 74.8 & 75.9 & 73.7 \\ 
 \hline
\end{tabular}}};
    \end{axis}
    
\end{tikzpicture}}
\caption{Precision-Recall evaluation for patch-level models in the DRIVE dataset.}
\label{fig:PR_values_patch_vessel}
\end{figure}

Figure~\ref{fig:PR_values_patch_vessel} shows the precision-recall curves, along with the best F measure obtained for each configuration.
The non-connectivity patch-level model, that is, the one that does not require
any learning about the connectivity reaches the best result (F=82.1).
The connectivity model, which has to tell apart those points connected with the patch center, achieves an only slightly worse performance (F=80.4), despite the task being more complicated.
The model that has also to distinguish between arteries and veins results in a more significant drop 
in the performance (F=74.8), but it still keeps a very good result.

\begin{figure}[t]
\centering
\includegraphics[width=\linewidth]{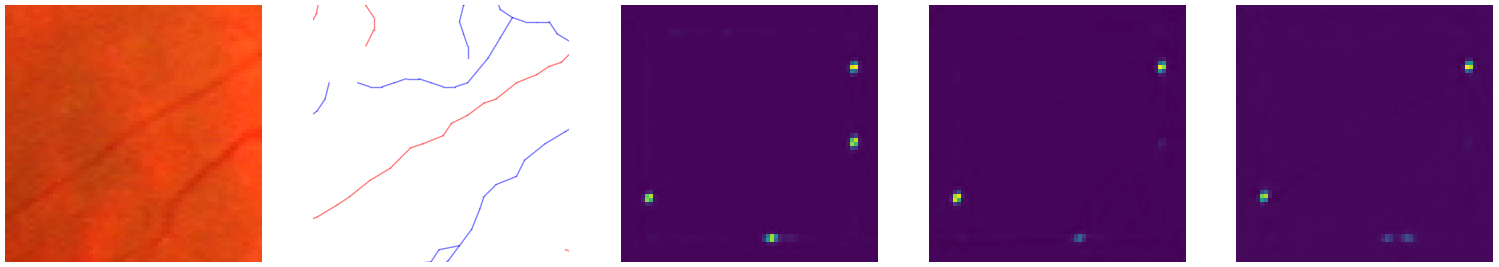}\\
\includegraphics[width=\linewidth]{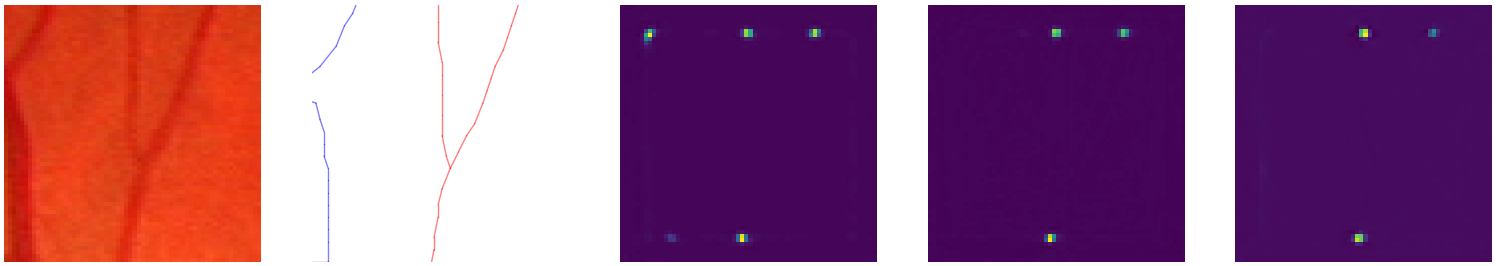}\\
\includegraphics[width=\linewidth]{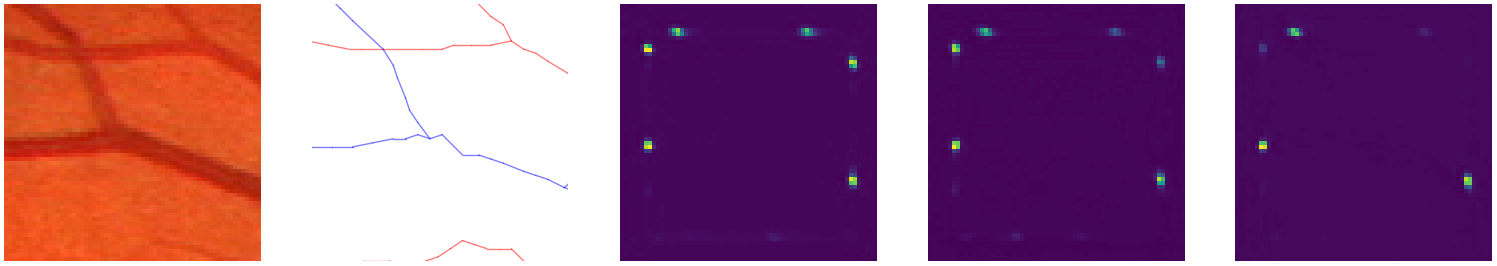}\\
\includegraphics[width=\linewidth]{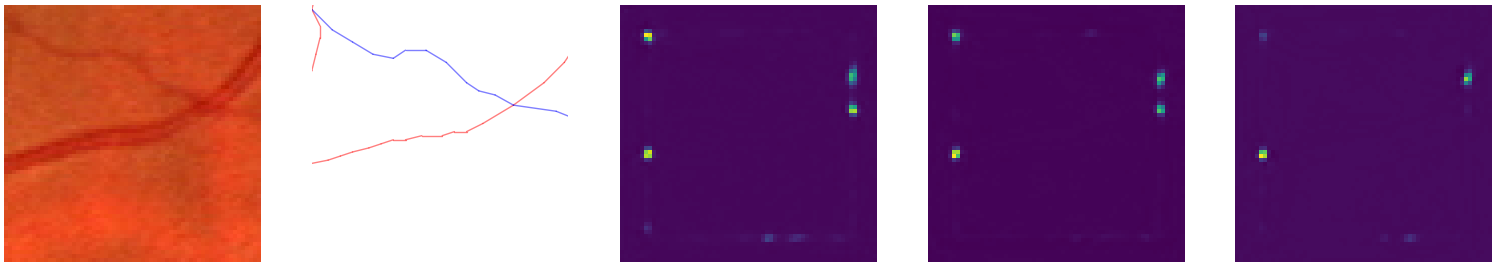}\\
\includegraphics[width=\linewidth]{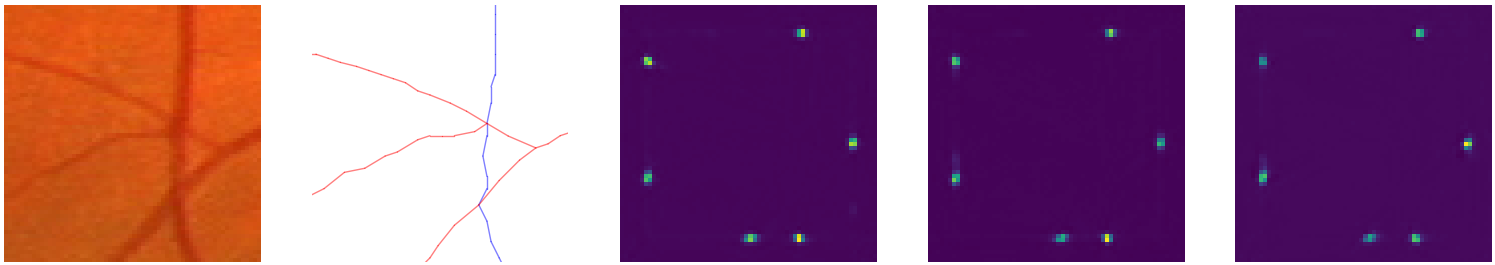}\\
\caption{Visual results of the patch-level models for eye fundus images. From left to right: eye fundus image, artery-vein annotation, output confidence for non-connectivity model, 
output confidence for connectivity model and output confidence for connectivity-av (artery-vein) model.}
\label{fig:vessel_patch_results}
\end{figure}

Figure~\ref{fig:vessel_patch_results} shows some visual results for the three type of configurations considered.
In the two first rows, the model is able to differentiate the vessels connected to the patch center from those ones not connected (3rd and 4th column). On the third and fourth rows, the model differentiates the vessels from the 
same type as the centered vessel (an artery) from those of different vessel type (see 4th and 5th column).
The last row shows a failure case where the model correctly predicts the connectivity but it is
not able to differentiate the arteries from the veins.

\paragraph{Iterative delineation:}
Once the patch-level model for connectivity has been trained, it is iteratively applied to extract the topology of the blood vessels networks from the eye fundus images.
As a strong baseline we compare to extracting the morphological skeleton of the DRIU~\cite{maninis2016deep} detections binarized at a certain threshold.
Table~\ref{tab:PR_values_global_vessel} compares to DRIU for different thresholds: 224 (the optimal for vessel segmentation obtained in \cite{maninis2016deep}), 200 (the optimal value for precision-recall boundary evaluation $F1^R$) and 170 (the optimal value for precision-connectivity evaluation $F1^C$).
Our proposed iterative approach outperforms DRIU for connectivity in 6.6 points, which results on a improvement of 1.8 in the precision-connectivity evaluation measure $F1^C$.
Furthermore, both techniques are also compared with an upper bound and a lower bound: the former is the skeleton extracted from the ground truth vessel segmentation, and the latter results from evaluating
the ground truth skeleton obtained from a different image.
Our results are only 7.7 points below the upper bound in connectivity.

\begin{table}
\resizebox{\linewidth}{!}{%
\begin{tabular}{ |c|c|c|c|c|c| } 
 \hline
 & $F1^R$ & P & R & C & $F1^C$\\ 
 \hline
 DRIU-224 \cite{maninis2016deep} & 90.4 & \bf{97.3} & 84.7 & 67.7 & 79.8\\
 DRIU-200 \cite{maninis2016deep} & \bf{92.0} & 93.8 & 90.6 & 74.0 & 82.7\\
 DRIU-170 \cite{maninis2016deep} & 91.3 & 89.9 & 93.1 & 78.3 & 83.7\\
 Iterative (ours) & 89.8 & 86.1 & \bf{94.1} & \bf{84.9} & \bf{85.5}\\
 \hline 
 GT skel (upperbound) & 97.4 & 95.6 & 99.3 & 92.6 & 94.1\\ 
 Random (lowerbound) & 44.9 & 44.2 & 45.9 & 21.8 & 29.2\\
 \hline
\end{tabular}}
\vspace{1mm}
\caption{Boundary Precision-Recall and Connectivity evaluation for vessels in DRIVE dataset.}
\label{tab:PR_values_global_vessel}
\end{table}

\begin{figure}[t]
\centering
\setlength{\fboxsep}{0pt}
\fbox{\includegraphics[width=0.32\linewidth]{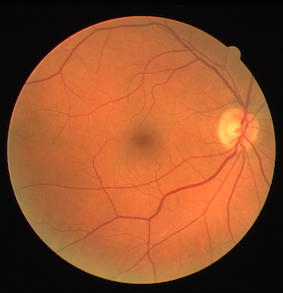}}\hfill
\fbox{\includegraphics[width=0.32\linewidth]{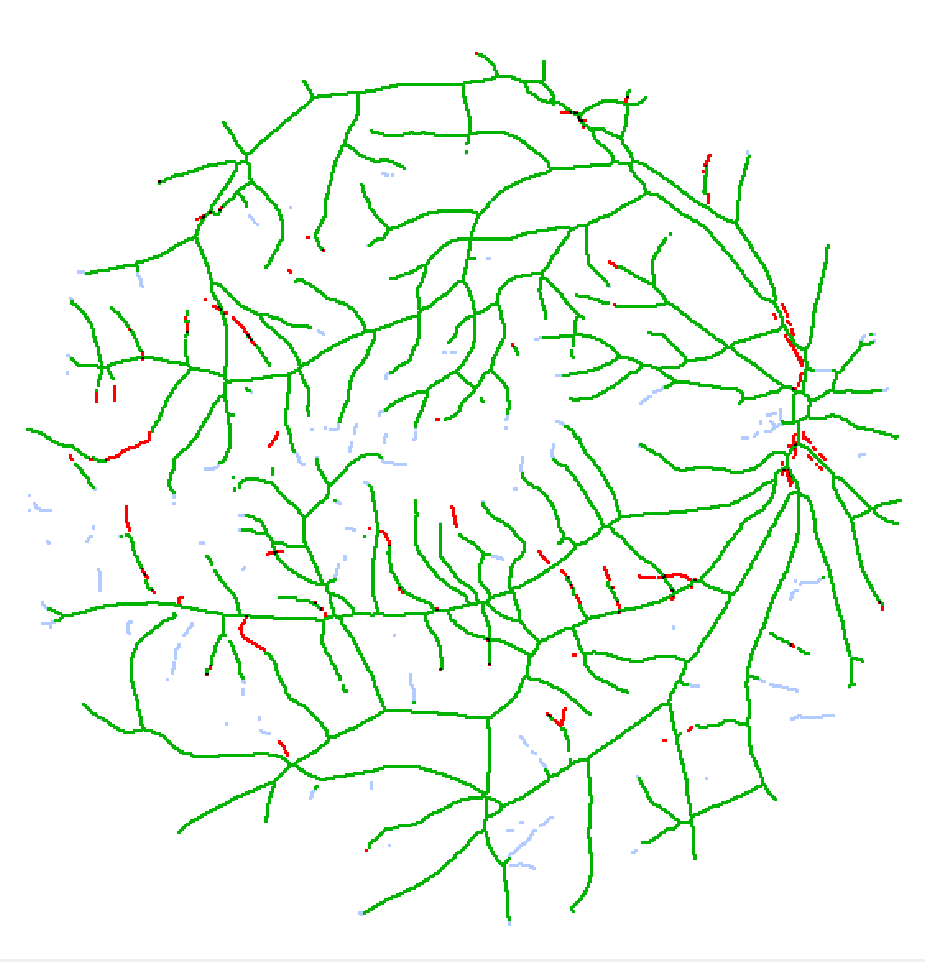}}\hfill
\fbox{\includegraphics[width=0.32\linewidth]{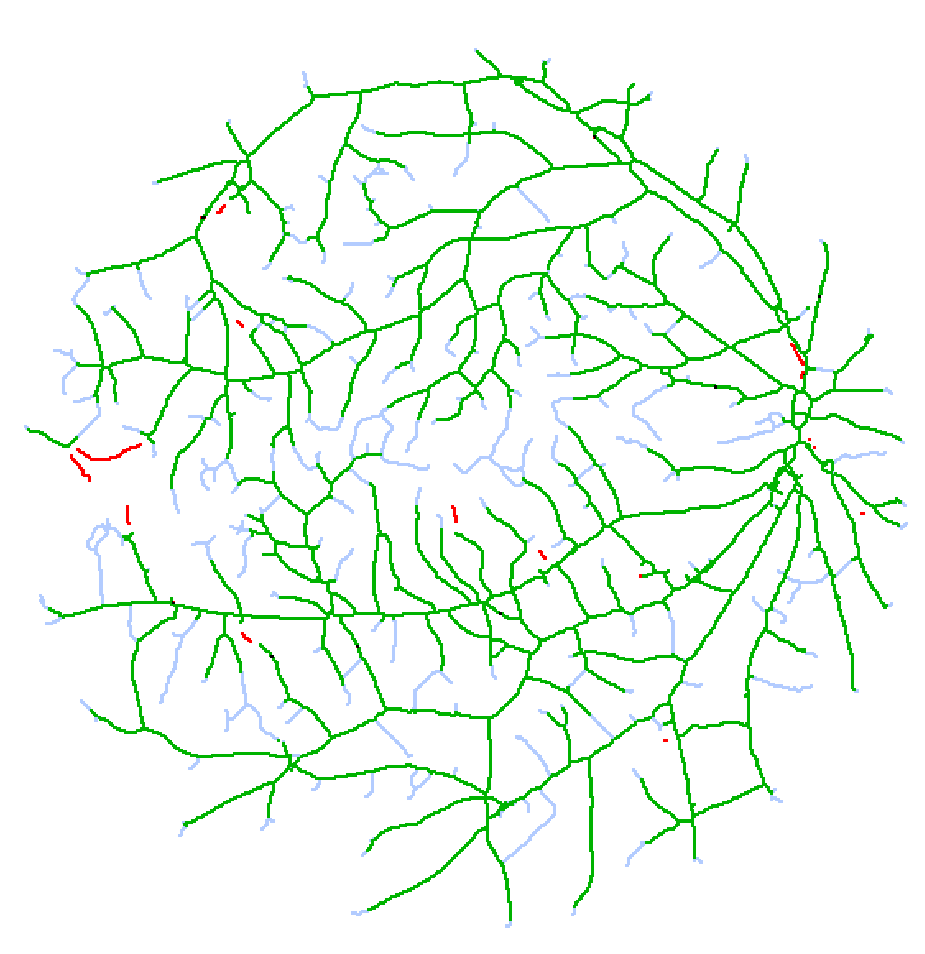}}\\[1mm]
\fbox{\includegraphics[width=0.32\linewidth]{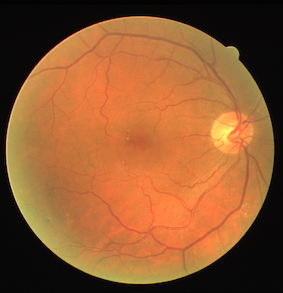}}\hfill
\fbox{\includegraphics[width=0.32\linewidth]{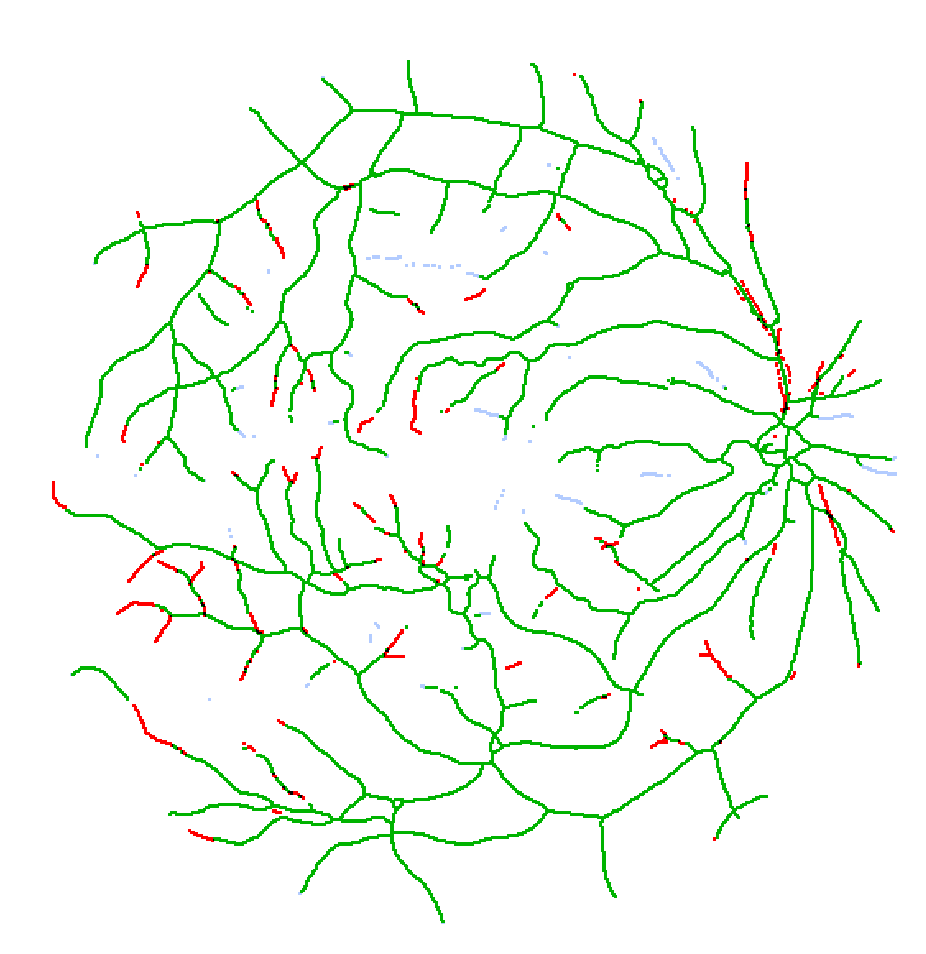}}\hfill
\fbox{\includegraphics[width=0.32\linewidth]{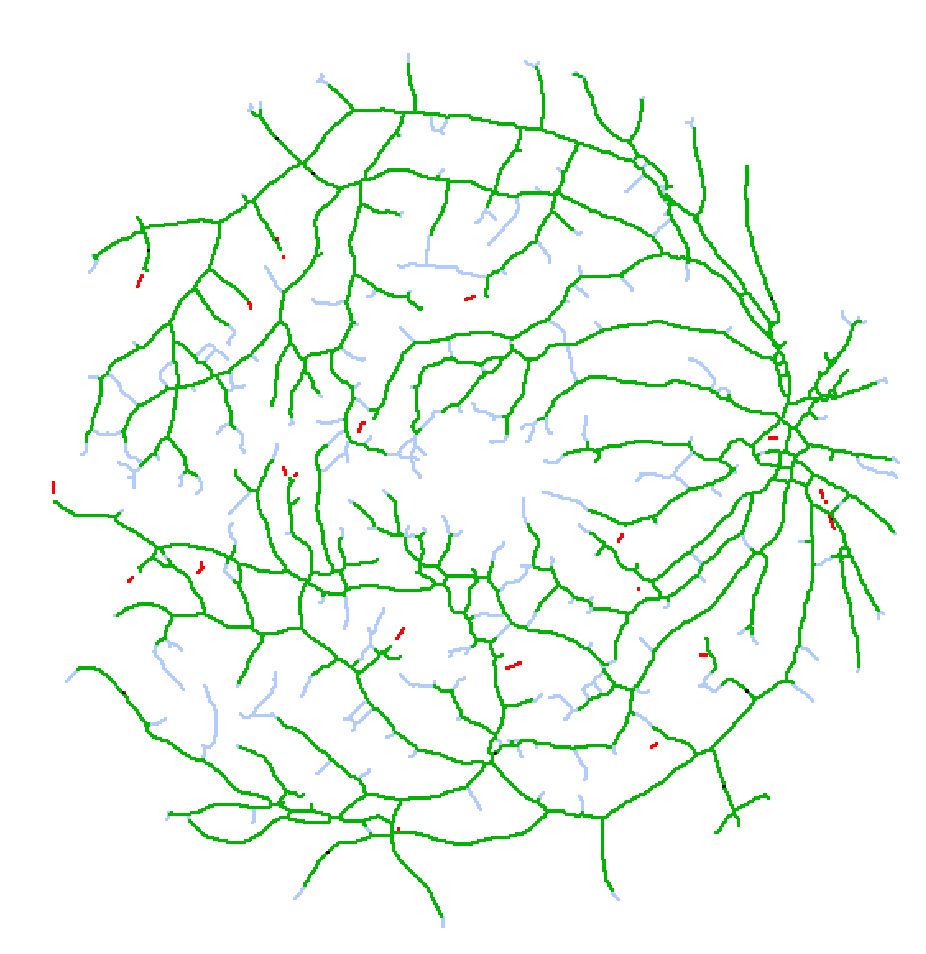}}\\
\caption{Qualitative results comparing DRIU (left) with our method (right): False positives in blue, false negatives in red.}
\label{fig:qualitative-results-vessels}
\end{figure}

Figure~\ref{fig:qualitative-results-vessels} shows some visual qualitative results, where the green pixels represent true positive contours, blue pixels represent false positives and red pixels false negatives.
We see that the main failure of our method are an over-extension of the ends of the vessels with respect to the annotation. In the majority of cases there is, albeit very weak, some trace of such vessels.
Out technique has less false negatives than DRIU, which have a higher impact in the connectivity.

%
%
%

\begin{figure}[t]
\centering
\setlength{\fboxsep}{0pt}
\adjincludegraphics[width=0.325\linewidth,trim={{.05\width} {.05\height} {.03\width} {.05\height}},clip]{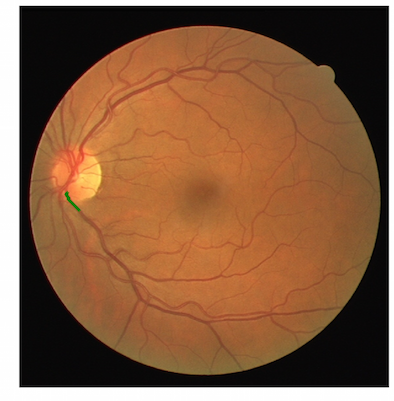}\hfill
\adjincludegraphics[width=0.325\linewidth,trim={{.05\width} {.05\height} {.03\width} {.05\height}},clip]{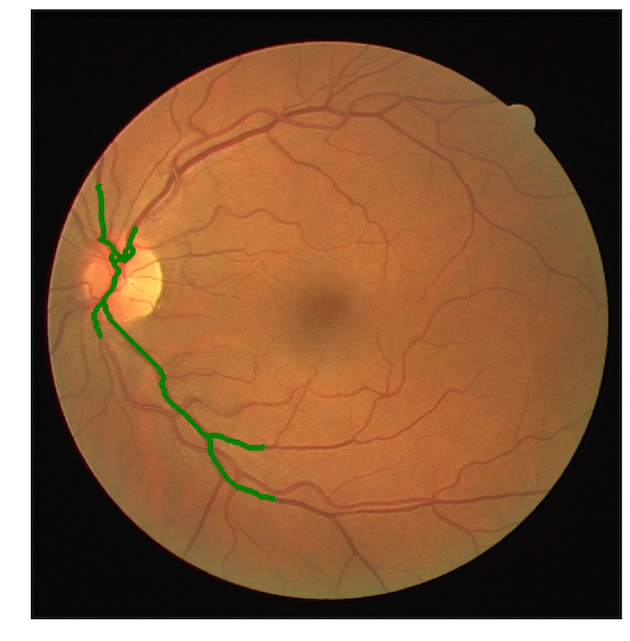}\hfill
\adjincludegraphics[width=0.325\linewidth,trim={{.05\width} {.05\height} {.03\width} {.05\height}},clip]{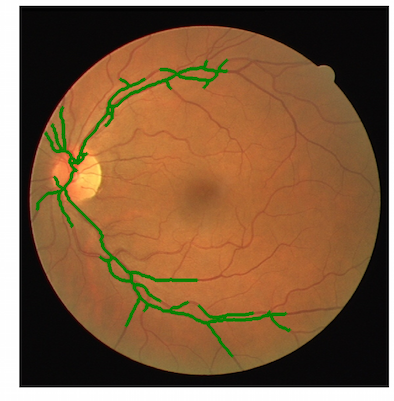}\\[1mm]
\adjincludegraphics[width=0.325\linewidth,trim={{.05\width} {.05\height} {.03\width} {.05\height}},clip]{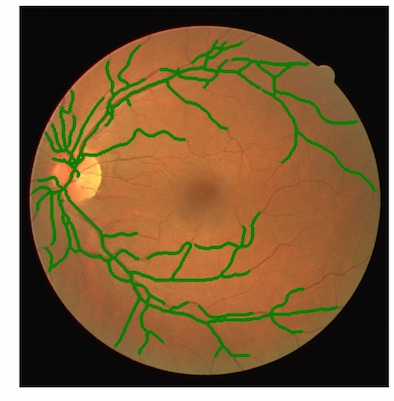}\hfill
\adjincludegraphics[width=0.325\linewidth,trim={{.05\width} {.05\height} {.03\width} {.05\height}},clip]{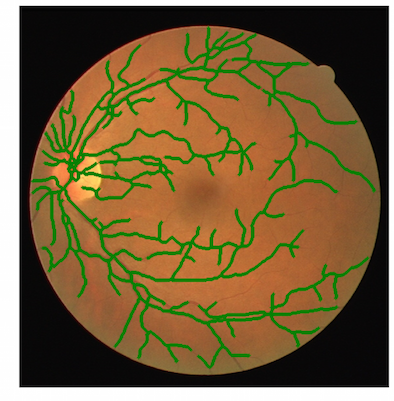}\hfill
\adjincludegraphics[width=0.325\linewidth,trim={{.05\width} {.05\height} {.03\width} {.05\height}},clip]{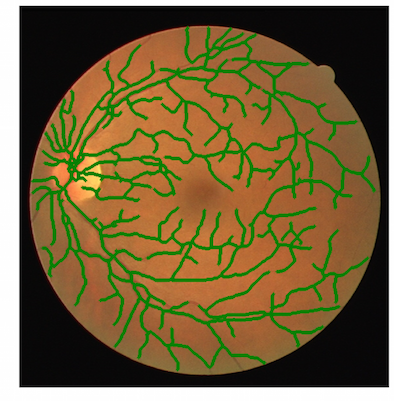}\\\caption{Evolution of the vessel network in the iterative delineation. Progress is displayed from left to right and from top to bottom.}
\label{fig:progress-iterative-approach}
\end{figure}

Figure~\ref{fig:progress-iterative-approach} illustrates how the vessel network topology extraction evolves along the iterations of our proposed approach for one of the test images.

\paragraph{Arteries and veins separation:}
For eye fundus images, we also pursue the objective of differentiating arteries and veins.
The approach is similar to the iterative delineation proposed before, but now using the patch-level model for connectivity that also takes into account that the vessels connected have to be of the
same type.
We have referred before to this model as the \emph{connectivity-av} model.

\begin{figure}[t]
\centering
\setlength{\fboxsep}{0pt}
\adjincludegraphics[height=0.325\linewidth]{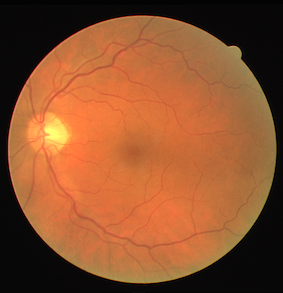}\hfill
\adjincludegraphics[height=0.325\linewidth]{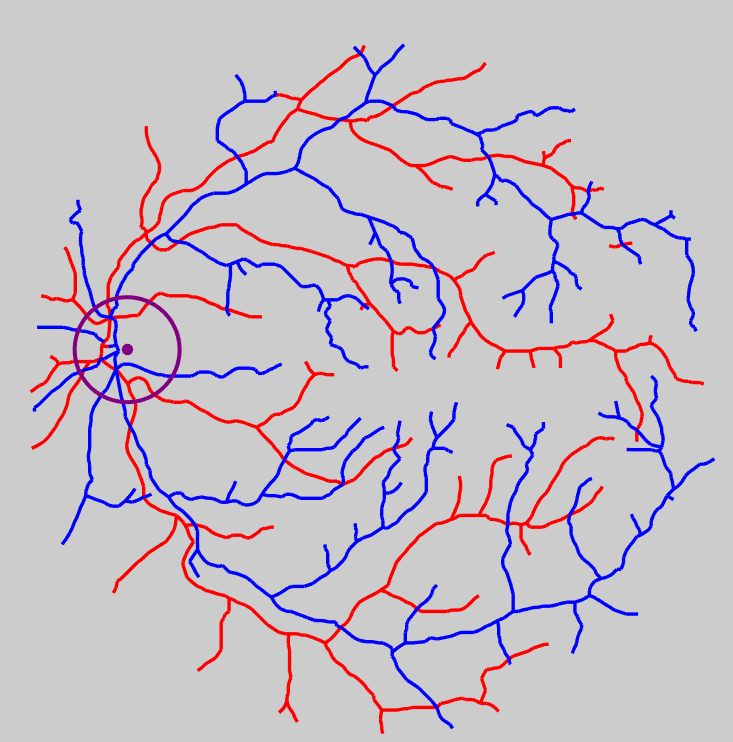}\hfill
\adjincludegraphics[height=0.325\linewidth]{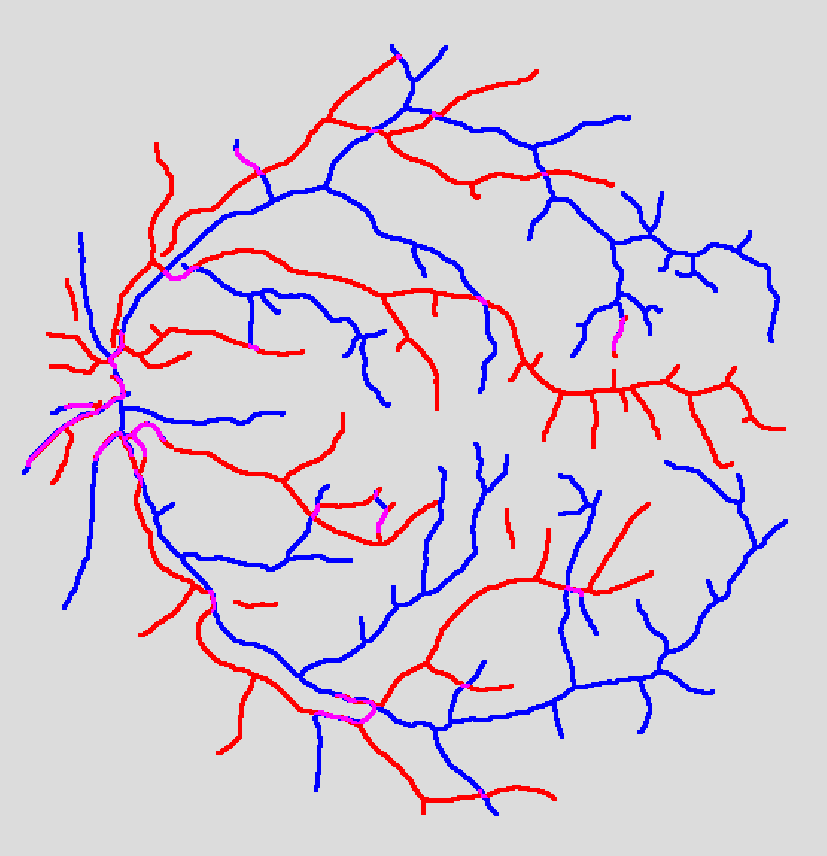}\\[1mm]
\adjincludegraphics[height=0.325\linewidth]{17_test.png}\hfill
\adjincludegraphics[height=0.325\linewidth]{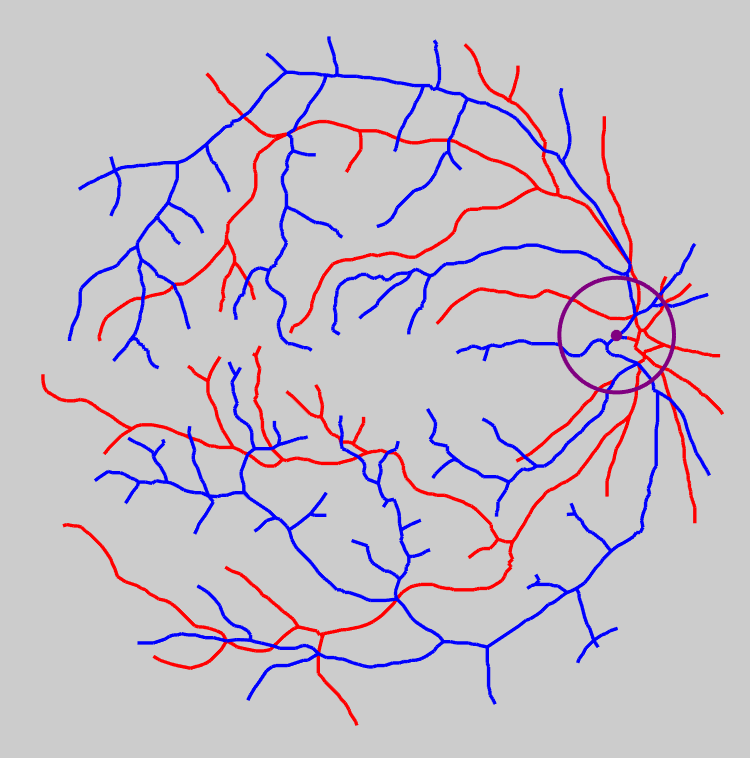}\hfill
\adjincludegraphics[height=0.325\linewidth]{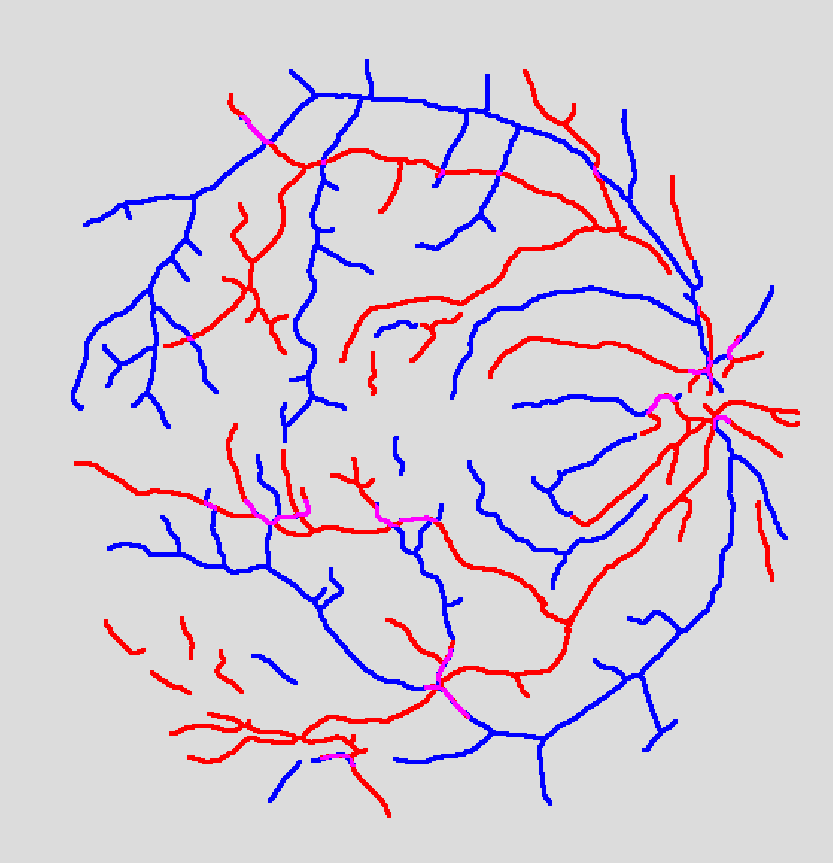}\\
\caption{Qualitative results on arteries and veins separation comparing ground truth (left) with our method (right): veins in blue, arteries in red.}
\label{fig:artery-vein-results}
\end{figure}

As baseline, we have considered the same CNN architecture as in DRIU, i.e.\ a VGG-based architecture, but using the annotations for arteries and veins given by~\cite{estrada2015retinal}.
These annotations are only given at the graph level, so we build the ground-truth image by drawing one-pixel wide lines delineating the arteries and veins networks; which is different from the vessel segmentation pixel-accurate masks from DRIVE on which DRIU is usually trained.
We train one global model for arteries and one for veins, and then we apply the delineation algorithm using the connectivity-av patch-level mode.
Table \ref{tab:artery-results} shows the results obtained for arteries (top) and veins (bottom).
In both cases, our iterative approach reaches the best trade off between $F1^R$ and $F1^C$. Figure~\ref{fig:artery-vein-results} shows some qualitative results comparing the ground truth
annotations with our method.

\begin{table}
\centering
\begin{tabular}{ |c|c|c|c|c|c| } 
 \hline
 & $F1^R$ & P & R & C & $F1^C$\\ 
 \hline
 VGG-220 & 76.1 & 72.9 & 80.7 & 52.4 & 61.0\\ 
 VGG-190 & 74.1 & 64.5 & \bf{88.2} & \bf{65.4} & 64.9\\ 
 Iterative (ours) & \bf{78.0} & \bf{81.4} & 75.3 & 63.0 & \bf{71.0}\\
 \hline
\end{tabular}\\[2mm]
\begin{tabular}{ |c|c|c|c|c|c| } 
 \hline
 & $F1^R$ & P & R & C & $F1^C$\\ 
 \hline
 VGG-230 & 74.2 & 70.8 & 79.1 & 42.2 & 52.9\\ 
 VGG-180 & 70.2 & 57.4 & \bf{91.3} & \bf{66.1} & 61.5\\
 Iterative (ours) & \bf{75.4} & \bf{72.0} & 79.6 & 61.2 & \bf{66.2}\\
 \hline
\end{tabular}
\vspace{1mm}
\caption{Boundary Precision-Recall and Connectivity evaluation for arteries (top) and veins (bottom) in the DRIVE dataset}
\label{tab:artery-results}
\end{table}


\subsection{Road topology on aerial images}
The experiments for road topology extraction on aerial images have been carried out on the Massachusetts Roads Dataset~\cite{MnihThesis}, which includes 1108 images for training, 14
images for validation, and 49 images for testing.

As done for the retinal images, we first train the patch-level model (in the case of roads, only for connectivity since there is only one type of road annotated).
We use the same patch size (64 $\times$ 64) but given that the resolution of the aerial images is significantly higher (1500 $\times$ 1500) than that of the retinal images (565 $\times$ 584),
130 patches have been randomly selected for each aerial image to train the model.
We obtain a precision value of 86.8, a recall value of 82.2 and F=84.5\%.



\begin{figure}[t]
\centering
\setlength{\fboxsep}{0pt}
\fbox{\adjincludegraphics[width=0.24\linewidth,trim={{0.1\width} {0.1\height} {0.05\width} {0.05\height}},clip]{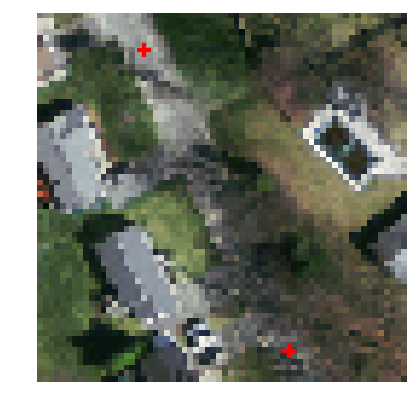}}\hfill
\fbox{\adjincludegraphics[width=0.24\linewidth,trim={{0.1\width} {0.1\height} {0.05\width} {0.05\height}},clip]{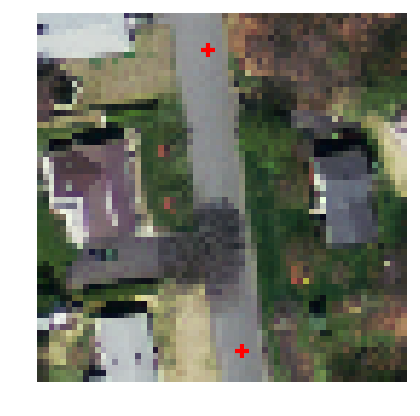}}\hfill
\fbox{\adjincludegraphics[width=0.24\linewidth,trim={{0.1\width} {0.1\height} {0.05\width} {0.05\height}},clip]{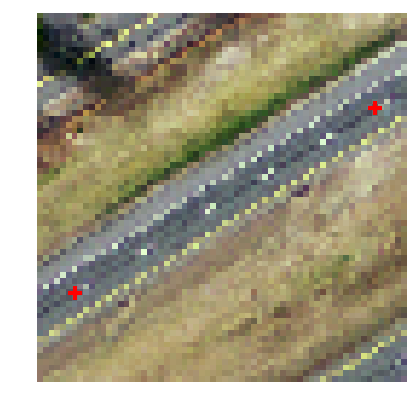}}\hfill
\fbox{\adjincludegraphics[width=0.24\linewidth,trim={{0.1\width} {0.1\height} {0.05\width} {0.05\height}},clip]{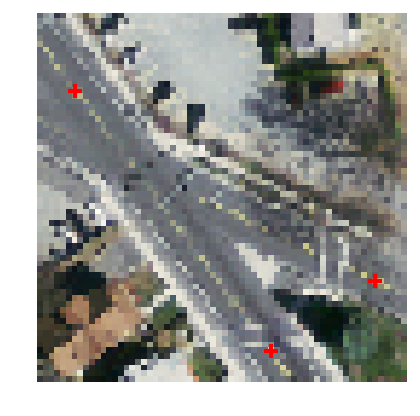}}\\[1pt]
\fbox{\adjincludegraphics[width=0.24\linewidth,trim={{0.1\width} {0.1\height} {0.05\width} {0.05\height}},clip]{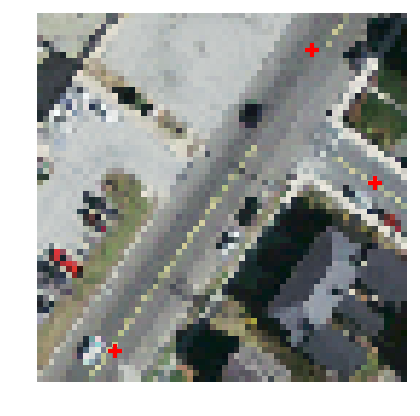}}\hfill
\fbox{\adjincludegraphics[width=0.24\linewidth,trim={{0.1\width} {0.1\height} {0.05\width} {0.05\height}},clip]{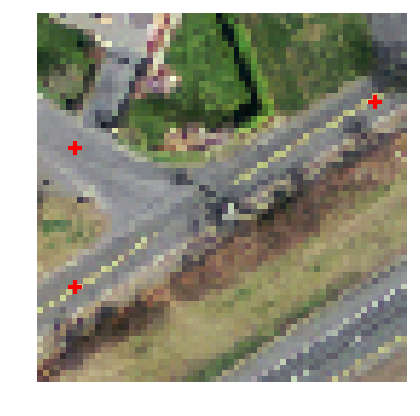}}\hfill
\fbox{\adjincludegraphics[width=0.24\linewidth,trim={{0.1\width} {0.1\height} {0.05\width} {0.05\height}},clip]{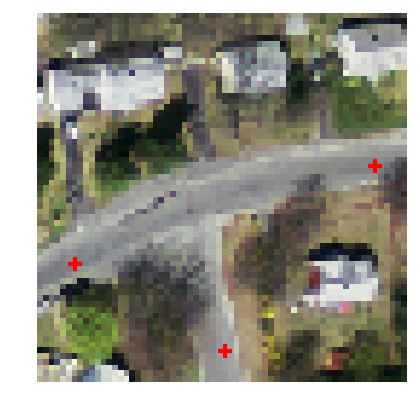}}\hfill
\fbox{\adjincludegraphics[width=0.24\linewidth,trim={{0.1\width} {0.1\height} {0.05\width} {0.05\height}},clip]{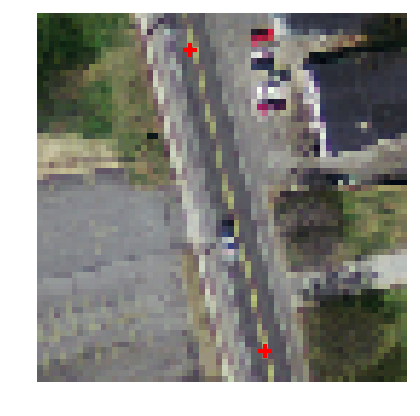}}\\[1pt]
\fbox{\adjincludegraphics[width=0.24\linewidth,trim={{0.1\width} {0.1\height} {0.05\width} {0.05\height}},clip]{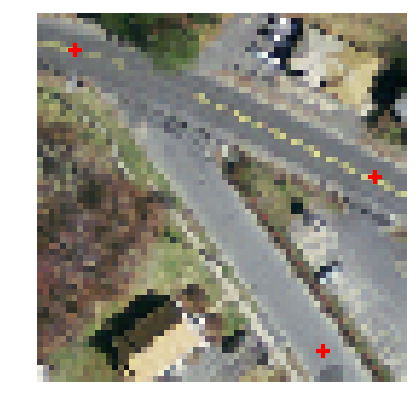}}\hfill
\fbox{\adjincludegraphics[width=0.24\linewidth,trim={{0.1\width} {0.1\height} {0.05\width} {0.05\height}},clip]{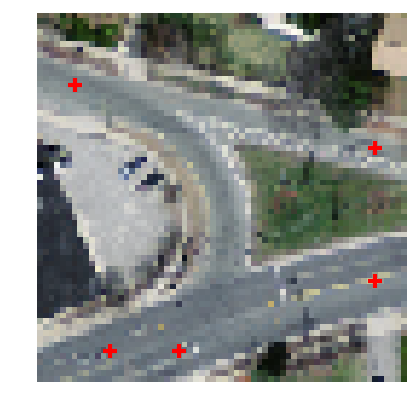}}\hfill
\fbox{\adjincludegraphics[width=0.24\linewidth,trim={{0.1\width} {0.1\height} {0.05\width} {0.05\height}},clip]{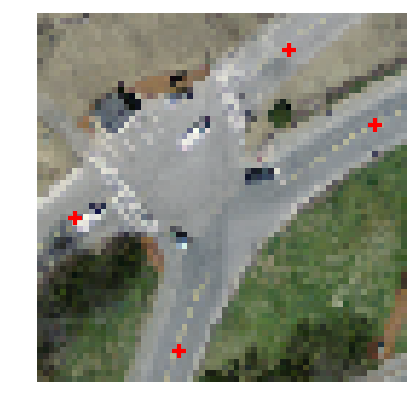}}\hfill
\fbox{\adjincludegraphics[width=0.24\linewidth,trim={{0.1\width} {0.1\height} {0.05\width} {0.05\height}},clip]{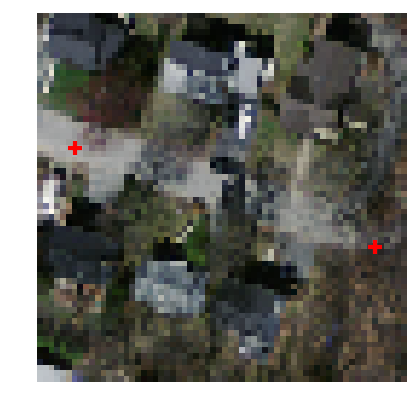}}\\[1pt]
\fbox{\adjincludegraphics[width=0.24\linewidth,trim={{0.1\width} {0.1\height} {0.05\width} {0.05\height}},clip]{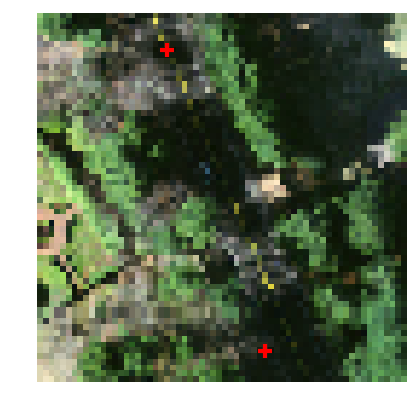}}\hfill
\fbox{\adjincludegraphics[width=0.24\linewidth,trim={{0.1\width} {0.1\height} {0.05\width} {0.05\height}},clip]{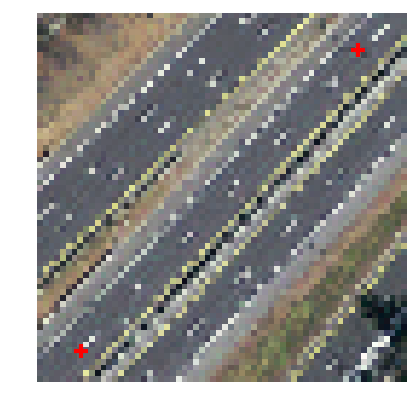}}\hfill
\fbox{\adjincludegraphics[width=0.24\linewidth,trim={{0.1\width} {0.1\height} {0.05\width} {0.05\height}},clip]{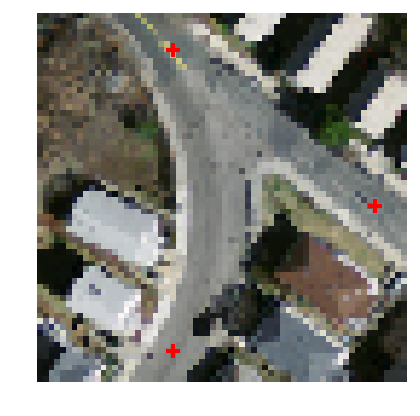}}\hfill
\fbox{\adjincludegraphics[width=0.24\linewidth,trim={{0.1\width} {0.1\height} {0.05\width} {0.05\height}},clip]{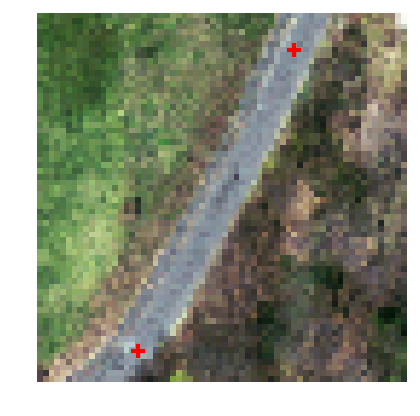}}\\
 
\caption{Visual results for the patch-level model for connectivity applied to patches from the Massachusetts Roads dataset. The red crosses represent the detections.}
\label{fig:patch-detector-roads}
\end{figure}

\begin{figure}[t]
\centering
\setlength{\fboxsep}{0pt}
\fbox{\adjincludegraphics[width=0.24\linewidth,trim={{0.1\width} {0.1\height} {0.05\width} {0.05\height}},clip]{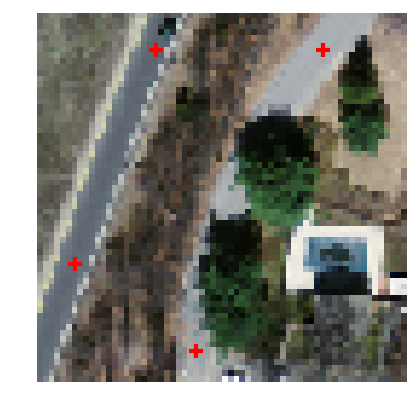}}\hfill
\fbox{\adjincludegraphics[width=0.24\linewidth,trim={{0.1\width} {0.1\height} {0.05\width} {0.05\height}},clip]{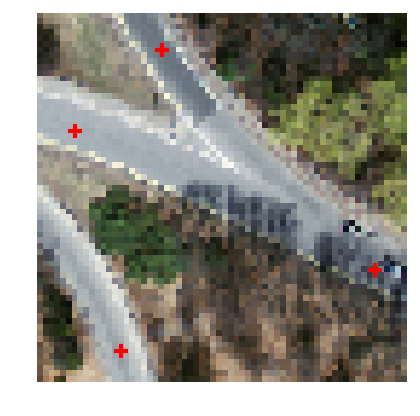}}\hfill
\fbox{\adjincludegraphics[width=0.24\linewidth,trim={{0.1\width} {0.1\height} {0.05\width} {0.05\height}},clip]{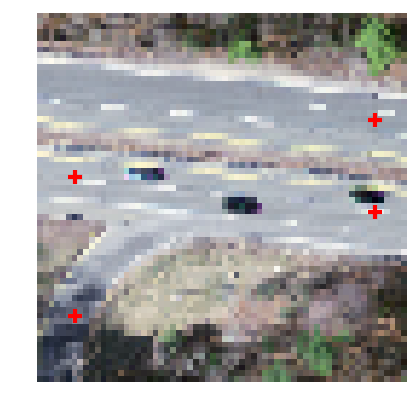}}\hfill
\fbox{\adjincludegraphics[width=0.24\linewidth,trim={{0.1\width} {0.1\height} {0.05\width} {0.05\height}},clip]{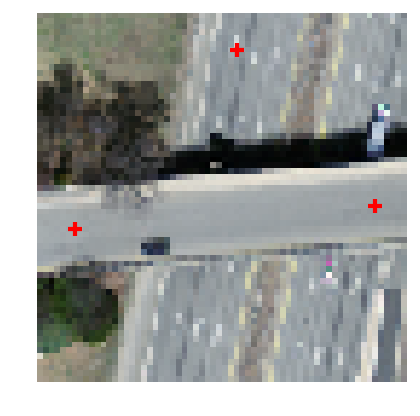}}\\[1mm]
\fbox{\adjincludegraphics[width=0.24\linewidth,trim={{0.1\width} {0.1\height} {0.05\width} {0.05\height}},clip]{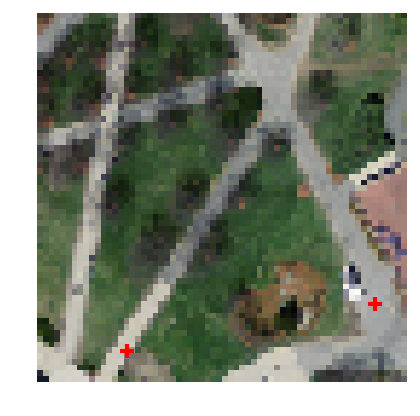}}\hfill
\fbox{\adjincludegraphics[width=0.24\linewidth,trim={{0.1\width} {0.1\height} {0.05\width} {0.05\height}},clip]{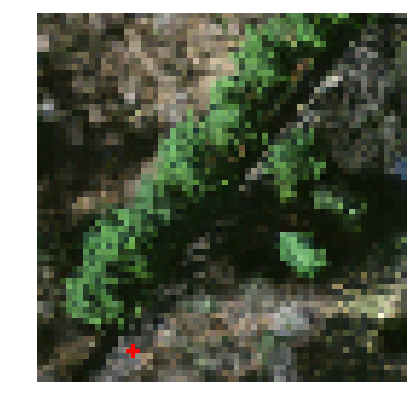}}\hfill
\fbox{\adjincludegraphics[width=0.24\linewidth,trim={{0.1\width} {0.1\height} {0.05\width} {0.05\height}},clip]{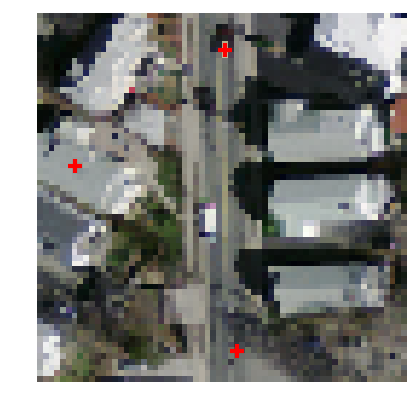}}\hfill
\fbox{\adjincludegraphics[width=0.24\linewidth,trim={{0.1\width} {0.1\height} {0.05\width} {0.05\height}},clip]{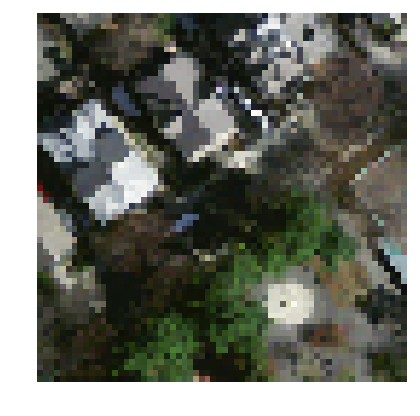}}\\
\caption{Detection errors for the patch-level model for connectivity applied to patches from the Massachusetts Roads dataset. The red crosses represent the detections.}
\label{fig:patch-detector-roads-errors}
\end{figure}

Figure~\ref{fig:patch-detector-roads} shows some results for the patch-level model for connectivity applied to patches from the Massachusetts Roads dataset.
We can see that there are some examples where the road connections are found despite the shadows of the trees or the similarity of the background with the road.
Furthermore, it also learns not to detect roads that are visible on the image but they are not connected to the central road.
Figure~\ref{fig:patch-detector-roads-errors} illustrates some other examples where the model fails with 
some false or missing detections. The three first images from the first row are examples where a visible road not connected with the central road has been wrongly detected. Other failure
example are missing detections on path roads, on hardly visible roads as well as false detection on building ceilings similar to roads.

Table~\ref{tab:roads-results} shows the comparison between the global-based baseline for road segmentation based on a VGG architecture and our proposed iterative approach.
The connectivity of the skeleton resulting from the VGG-based road segmentation is very low, with a maximum precision-connectivity value $F1^C = 49.3$.
This result is obtained with low values for both precision (49.2) and connectivity (49.4). 
In contrast, our iterative delineation approach is able to reach a maximum precision-connectivity value $F1^C = 74.4\%$.  

\begin{table}
\begin{tabular}{ |c|c|c|c|c|c| } 
 \hline
 & $F1^R$ & P & R & C & $F1^C$\\ 
 \hline
 VGG-150 & 64.1 & 49.2 & \bf{94.7} & 49.4 & 49.3\\
 VGG-175 & 72.0 & 61.5 & 88.6 & 30.7 & 41.0\\
 VGG-200 & 72.4 & 75.8 & 70.7 & 11.0 & 19.2\\
 Iterative-20 (ours) & 78.2 & 72.0 & 87.0 & \bf{73.4} & 72.7 \\
 Iterative-25 (ours) & 80.9 & 79.1 & 83.9 & 69.9 & 74.2 \\
 Iterative-30 (ours) & \bf{81.6} & \bf{83.5} & 80.8 & 67.1 & \bf{74.4} \\
 \hline
\end{tabular}
\vspace{1pt}
\caption{Boundary Precision-Recall and Connectivity evaluation in Massachusetts Roads dataset. VGG-150 refers to the threshold used on the output segmentation road from the VGG model before
extracting the skeleton. Iterative-20 refers to the threshold used on the patch-level model for connectivity in our iterative approach.}
\label{tab:roads-results}
\end{table}

\begin{figure}
\centering
\setlength{\fboxsep}{0pt}
\fbox{\adjincludegraphics[width=0.24\linewidth,trim={{.06\width} {.05\height} {.02\width} {.02\height}},clip]{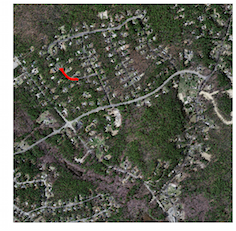}}\hfill
\fbox{\adjincludegraphics[width=0.24\linewidth,trim={{.06\width} {.05\height} {.02\width} {.02\height}},clip]{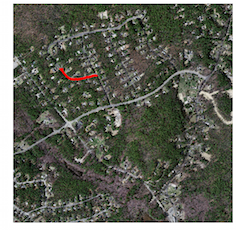}}\hfill
\fbox{\adjincludegraphics[width=0.24\linewidth,trim={{.06\width} {.05\height} {.02\width} {.02\height}},clip]{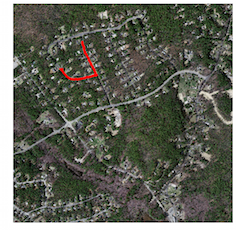}}\hfill
\fbox{\adjincludegraphics[width=0.24\linewidth,trim={{.06\width} {.05\height} {.02\width} {.02\height}},clip]{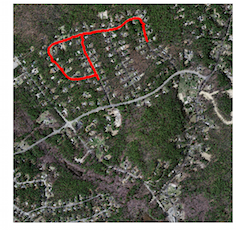}}\\[2pt]
\fbox{\adjincludegraphics[width=0.24\linewidth,trim={{.06\width} {.05\height} {.02\width} {.02\height}},clip]{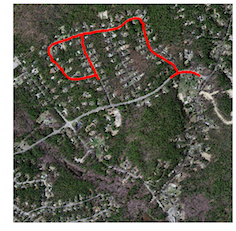}}\hfill
\fbox{\adjincludegraphics[width=0.24\linewidth,trim={{.06\width} {.05\height} {.02\width} {.02\height}},clip]{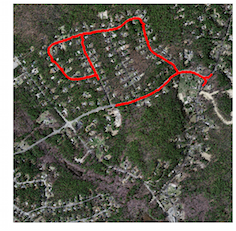}}\hfill
\fbox{\adjincludegraphics[width=0.24\linewidth,trim={{.06\width} {.05\height} {.02\width} {.02\height}},clip]{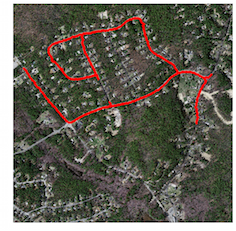}}\hfill
\fbox{\adjincludegraphics[width=0.24\linewidth,trim={{.06\width} {.05\height} {.02\width} {.02\height}},clip]{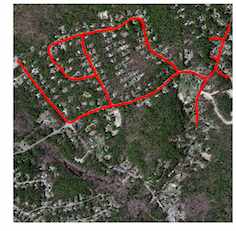}}\\[2pt]
\fbox{\adjincludegraphics[width=0.24\linewidth,trim={{.06\width} {.05\height} {.02\width} {.02\height}},clip]{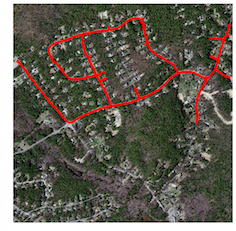}}\hfill
\fbox{\adjincludegraphics[width=0.24\linewidth,trim={{.06\width} {.05\height} {.02\width} {.02\height}},clip]{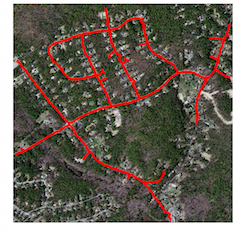}}\hfill
\fbox{\adjincludegraphics[width=0.24\linewidth,trim={{.06\width} {.05\height} {.02\width} {.02\height}},clip]{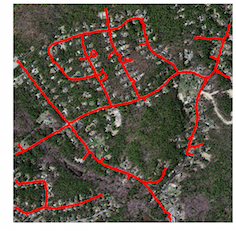}}\hfill
\fbox{\adjincludegraphics[width=0.24\linewidth,trim={{.06\width} {.09\height} {.02\width} {.02\height}},clip]{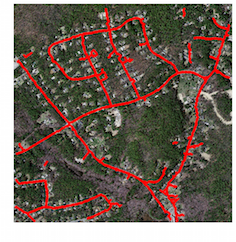}}\\
\caption{Evolution of the road network in the iterative delineation. The progress is displayed from left to right and from top to bottom.}
\label{fig:progress-iterative-approach-roads}
\end{figure}

\begin{figure}[t]
\centering
\setlength{\fboxsep}{0pt}
\begin{minipage}{0.33\linewidth}\footnotesize\centering VGG-150\end{minipage}\hfill
\begin{minipage}{0.33\linewidth}\footnotesize\centering Ours\end{minipage}\hfill
\begin{minipage}{0.33\linewidth}\footnotesize\centering Ground Truth\vphantom{p}\end{minipage}\\
\fbox{\adjincludegraphics[height=0.33\linewidth,trim={{0.12\width} {0.1\height} {0.1\width} {0.1\height}},clip]{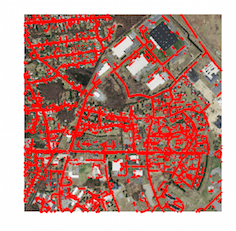}}\hfill
\fbox{\adjincludegraphics[height=0.33\linewidth,trim={{0.12\width} {0.1\height} {0.1\width} {0.1\height}},clip]{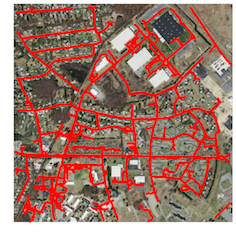}}\hfill
\fbox{\adjincludegraphics[height=0.33\linewidth,trim={{0.12\width} {0.1\height} {0.1\width} {0.1\height}},clip]{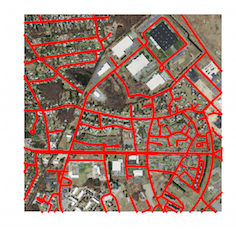}}\\[1pt]
\fbox{\adjincludegraphics[height=0.33\linewidth,trim={{0.12\width} {0.1\height} {0.1\width} {0.1\height}},clip]{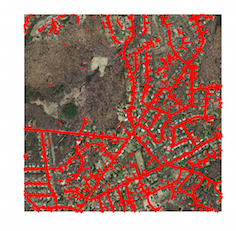}}\hfill
\fbox{\adjincludegraphics[height=0.33\linewidth,trim={{0.12\width} {0.1\height} {0.1\width} {0.1\height}},clip]{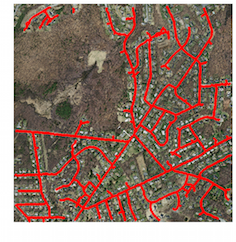}}\hfill
\fbox{\adjincludegraphics[height=0.33\linewidth,trim={{0.12\width} {0.1\height} {0.1\width} {0.1\height}},clip]{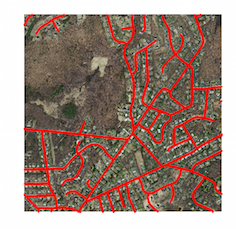}}\\[1pt]
\fbox{\adjincludegraphics[height=0.33\linewidth,trim={{0.12\width} {0.1\height} {0.1\width} {0.1\height}},clip]{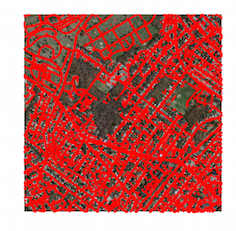}}\hfill
\fbox{\adjincludegraphics[height=0.33\linewidth,trim={{0.12\width} {0.1\height} {0.1\width} {0.1\height}},clip]{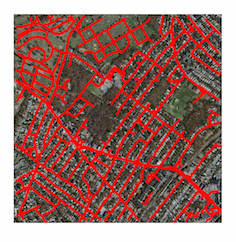}}\hfill
\fbox{\adjincludegraphics[height=0.33\linewidth,trim={{0.12\width} {0.1\height} {0.1\width} {0.1\height}},clip]{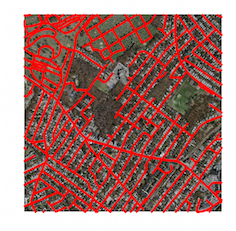}}\\[1pt]
\fbox{\adjincludegraphics[height=0.33\linewidth,trim={{0.12\width} {0.1\height} {0.1\width} {0.1\height}},clip]{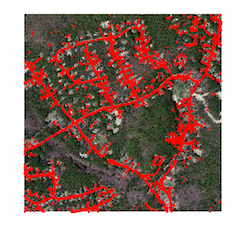}}\hfill
\fbox{\adjincludegraphics[height=0.33\linewidth,trim={{0.12\width} {0.1\height} {0.1\width} {0.1\height}},clip]{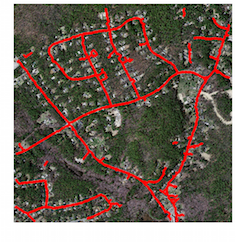}}\hfill
\fbox{\adjincludegraphics[height=0.33\linewidth,trim={{0.12\width} {0.1\height} {0.1\width} {0.1\height}},clip]{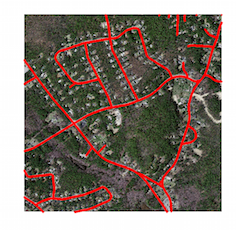}}\\
 
\caption{Results for road network topology extraction on the Massachusetts Roads dataset.
From left to right: VGG-150, our iterative delineation approach and ground truth.}
\label{fig:roads-results}
\end{figure}

\begin{figure}[t]
\centering
\setlength{\fboxsep}{0pt}
\fbox{\adjincludegraphics[height=0.32\linewidth,trim={{0.12\width} {0.1\height} {0.1\width} {0.1\height}},clip]{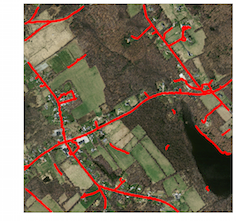}}\hfill
\fbox{\adjincludegraphics[height=0.32\linewidth,trim={{0.12\width} {0.1\height} {0.1\width} {0.1\height}},clip]{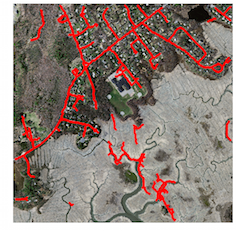}}\hfill
\fbox{\adjincludegraphics[height=0.32\linewidth,trim={{0.12\width} {0.1\height} {0.1\width} {0.1\height}},clip]{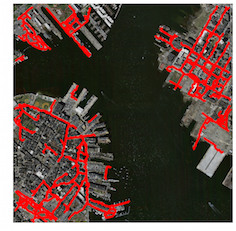}}\\
 
\caption{False and missing detections for road network topology extraction on the Massachusetts Roads dataset.}
\label{fig:roads-results-errors}
\end{figure}

Figure~\ref{fig:progress-iterative-approach-roads} illustrates how the road network topology delineation evolves along the iterations of our proposed approach for one of the test images.
Figure~\ref{fig:roads-results} shows some qualitative results in comparison with the VGG-based road segmentation baseline and the ground-truth annotations. Figure~\ref{fig:roads-results-errors}
shows some errors in the network topology delineation as some false detections on field (left image) and fluvial (central image) areas or missing detections on high density urban areas (right
image).

\section{Conclusions}

In this paper we have presented an approach that iteratively applies a patch-based CNN model for connectivity to extract the topology of filamentary networks.
We have demonstrated the effectiveness of our technique on retinal vessels from fundus images and road networks from aerial photos.
The patch-based model is capable of learning that the central point is
the input location and of finding the locations at the patch border connected to the center.
Furthermore, on the retinal images, we can also differentiate arteries and veins and extract their respective networks.
A new  F measure ($F1^C$) that combines precision and connectivity has been proposed to evaluate the topology results.  The experiments carried out on both retinal and aerial images have obtained the best performance on $F1^C$ compared to strong baselines.

{\small
\bibliographystyle{ieee}
\bibliography{cvpr18_biblio}

\begin{thebibliography}{10}\itemsep=-1pt

\bibitem{Ban+12}
P.~Bankhead, C.~N. Scholfield, J.~G. McGeown, and T.~M. Curtis.
\newblock Fast retinal vessel detection and measurement using wavelets and edge
  location refinement.
\newblock {\em PloS one}, 2012.

\bibitem{Bec+13}
C.~Becker, R.~Rigamonti, V.~Lepetit, and P.~Fua.
\newblock Supervised feature learning for curvilinear structure segmentation.
\newblock In {\em MICCAI}, 2013.

\bibitem{cheng2014tracing}
L.~Cheng, J.~De, X.~Zhang, F.~Lin, and H.~Li.
\newblock Tracing retinal blood vessels by matrix-forest theorem of directed
  graphs.
\newblock In {\em MICCAI}, 2014.

\bibitem{DIJKSTRA1959}
E.~Dijkstra.
\newblock A note on two problems in connexion with graphs.
\newblock {\em Numerische Mathematik}, 1:269--271, 1959.

\bibitem{estrada2015retinal}
R.~Estrada, M.~J. Allingham, P.~S. Mettu, S.~W. Cousins, C.~Tomasi, and
  S.~Farsiu.
\newblock Retinal artery-vein classification via topology estimation.
\newblock {\em IEEE transactions on medical imaging}, 34(12):2518--2534, 2015.

\bibitem{Fu+16}
H.~Fu, Y.~Xu, S.~Lin, D.~W.~K. Wong, and J.~Liu.
\newblock Deepvessel: Retinal vessel segmentation via deep learning and
  conditional random field.
\newblock In {\em MICCAI}, 2016.

\bibitem{Gir15}
R.~Girshick.
\newblock Fast {R-CNN}.
\newblock In {\em ICCV}, 2015.

\bibitem{Gir+14}
R.~Girshick, J.~Donahue, T.~Darrell, and J.~Malik.
\newblock Rich feature hierarchies for accurate object detection and semantic
  segmentation.
\newblock In {\em CVPR}, 2014.

\bibitem{GuCh15}
L.~Gu and L.~Cheng.
\newblock Learning to boost filamentary structure segmentation.
\newblock In {\em ICCV}, 2015.

\bibitem{He+17}
K.~He, G.~Gkioxari, P.~Doll{\'a}r, and R.~Girshick.
\newblock Mask {R-CNN}.
\newblock In {\em ICCV}, 2017.

\bibitem{He+16}
K.~He, X.~Zhang, S.~Ren, and J.~Sun.
\newblock Deep residual learning for image recognition.
\newblock In {\em CVPR}, 2016.

\bibitem{Krizhevsky2012}
A.~Krizhevsky, I.~Sutskever, and G.~E. Hinton.
\newblock Imagenet classification with deep convolutional neural networks.
\newblock In {\em NIPS}, 2012.

\bibitem{LaCh08}
M.~W. Law and A.~C. Chung.
\newblock Three dimensional curvilinear structure detection using optimally
  oriented flux.
\newblock In {\em ECCV}, pages 368--382. Springer, 2008.

\bibitem{Lee+17}
C.-Y. Lee, V.~Badrinarayanan, T.~Malisiewicz, and A.~Rabinovich.
\newblock Roomnet: End-to-end room layout estimation.
\newblock In {\em ICCV}, 2017.

\bibitem{Li+17}
Y.~Li, H.~Qi, J.~Dai, X.~Ji, and Y.~Wei.
\newblock Fully convolutional instance-aware semantic segmentation.
\newblock In {\em CVPR}, 2017.

\bibitem{LSD15}
J.~Long, E.~Shelhamer, and T.~Darrell.
\newblock Fully convolutional networks for semantic segmentation.
\newblock In {\em CVPR}, 2015.

\bibitem{maninis2016deep}
K.-K. Maninis, J.~Pont-Tuset, P.~Arbel{\'a}ez, and L.~Van~Gool.
\newblock Deep retinal image understanding.
\newblock In {\em International Conference on Medical Image Computing and
  Computer-Assisted Intervention}, pages 140--148. Springer, 2016.

\bibitem{martin2004learning}
D.~R. Martin, C.~C. Fowlkes, and J.~Malik.
\newblock Learning to detect natural image boundaries using local brightness,
  color, and texture cues.
\newblock {\em IEEE transactions on pattern analysis and machine intelligence},
  26(5):530--549, 2004.

\bibitem{mattyus2017deep}
G.~M{\'a}ttyus, W.~Luo, and R.~Urtasun.
\newblock Deeproadmapper: Extracting road topology from aerial images.
\newblock In {\em International Conference on Computer Vision}, 2017.

\bibitem{Mer+16}
J.~Merkow, A.~Marsden, D.~Kriegman, and Z.~Tu.
\newblock Dense volume-to-volume vascular boundary detection.
\newblock In {\em MICCAI}, 2016.

\bibitem{MnihThesis}
V.~Mnih.
\newblock {\em Machine Learning for Aerial Image Labeling}.
\newblock PhD thesis, University of Toronto, 2013.

\bibitem{NYD16}
A.~Newell, K.~Yang, and J.~Deng.
\newblock Stacked hourglass networks for human pose estimation.
\newblock In {\em ECCV}, 2016.

\bibitem{OrBl14}
J.~I. Orlando and M.~Blaschko.
\newblock Learning fully-connected crfs for blood vessel segmentation in
  retinal images.
\newblock In {\em MICCAI}, 2014.

\bibitem{Ren+15}
S.~Ren, K.~He, R.~Girshick, and J.~Sun.
\newblock Faster {R-CNN}: Towards real-time object detection with region
  proposal networks.
\newblock In {\em NIPS}, 2015.

\bibitem{RiPe07}
E.~Ricci and R.~Perfetti.
\newblock Retinal blood vessel segmentation using line operators and support
  vector classification.
\newblock {\em IEEE Transactions on Medical Imaging}, 26(10):1357--1365, 2007.

\bibitem{Rus+15}
O.~Russakovsky, J.~Deng, H.~Su, J.~Krause, S.~Satheesh, S.~Ma, Z.~Huang,
  A.~Karpathy, A.~Khosla, M.~Bernstein, A.~C. Berg, and L.~Fei-Fei.
\newblock {ImageNet Large Scale Visual Recognition Challenge}.
\newblock {\em IJCV}, 2015.

\bibitem{SiZi15}
K.~Simonyan and A.~Zisserman.
\newblock Very deep convolutional networks for large-scale image recognition.
\newblock In {\em ICLR}, 2015.

\bibitem{SLF14}
A.~Sironi, V.~Lepetit, and P.~Fua.
\newblock Multiscale centerline detection by learning a scale-space distance
  transform.
\newblock In {\em CVPR}, 2014.

\bibitem{SLF15}
A.~Sironi, V.~Lepetit, and P.~Fua.
\newblock Projection onto the manifold of elongated structures for accurate
  extraction.
\newblock In {\em ICCV}, 2015.

\bibitem{Soa+06}
J.~V. Soares, J.~J. Leandro, R.~M. Cesar~Jr, H.~F. Jelinek, and M.~J. Cree.
\newblock Retinal vessel segmentation using the 2-d gabor wavelet and
  supervised classification.
\newblock {\em IEEE Transactions on Medical Imaging}, 25(9):1214--1222, 2006.

\bibitem{staal04ridge}
J.~Staal, M.~Abramoff, M.~Niemeijer, M.~Viergever, and B.~van Ginneken.
\newblock {Ridge based vessel segmentation in color images of the retina}.
\newblock {\em {IEEE Transactions on Medical Imaging}}, 23(4):501--509, 2004.

\bibitem{wang2017torontocity}
S.~Wang, M.~Bai, G.~M{\'{a}}ttyus, H.~Chu, W.~Luo, B.~Yang, J.~Liang,
  J.~Cheverie, S.~Fidler, and R.~Urtasun.
\newblock Torontocity: Seeing the world with a million eyes.
\newblock In {\em International Conference on Computer Vision}, 2017.

\bibitem{WAC11}
Y.~Wang, A.~Narayanaswamy, C.-L. Tsai, and B.~Roysam.
\newblock A broadly applicable 3-d neuron tracing method based on open-curve
  snake.
\newblock {\em Neuroinformatics}, 9(2-3):193--217, 2011.

\bibitem{WMS13}
J.~D. Wegner, J.~A. Montoya-Zegarra, and K.~Schindler.
\newblock A higher-order crf model for road network extraction.
\newblock In {\em CVPR}, 2013.

\bibitem{XiTu17}
S.~Xie and Z.~Tu.
\newblock Holistically-nested edge detection.
\newblock {\em IJCV}, pages 1--16, 2017.

\end{thebibliography}
}

\end{document}